\documentclass[11pt,journal,letterpaper]{IEEEtran}
\usepackage{calc}
\usepackage{graphicx}
\usepackage{amsmath,amssymb} 
\usepackage{color}
\usepackage{subfigure}
\usepackage{cite}
\usepackage{booktabs}
\usepackage{xspace}
\newcommand{\subparagraph}{}
\usepackage[small,compact]{titlesec}
\usepackage[justification=justified,singlelinecheck=false]{caption}

\def\etal{\emph{et al.}\xspace}
\ifCLASSOPTIONcompsoc
\else
\fi

\ifCLASSINFOpdf
\else
\fi
\begin{document}

\title{Covariance of Motion and Appearance Features \\ 
for Human Action and Gesture Recognition}

\author{Subhabrata~Bhattacharya,
        Nasim~Souly
        and~Mubarak~Shah
\thanks{The authors are with the College of Electrical Engineering and Computer
Science, University of Central Florida, Orlando, FL, 32826
e-mail: (subh@cs.ucf.edu, souly@cs.ucf.edu, shah@cs.ucf.edu).}}

\IEEEcompsoctitleabstractindextext{
\begin{abstract}
In this paper, we introduce a novel descriptor for employing covariance of 
motion and appearance features for human action and gesture recognition. In 
our approach, we compute kinematic features from optical flow and 
first and second-order derivatives of intensities to represent motion and 
appearance respectively. These features are then used to construct covariance 
matrices which capture joint statistics of both low-level motion and 
appearance features extracted from a video. Using an over-complete dictionary 
of the covariance based descriptors built from labeled training samples, we 
formulate human action recognition as a sparse linear approximation problem. 
Within this, we pose the sparse decomposition of a covariance matrix, which 
also conforms to the space of semi-positive definite matrices, as a 
determinant maximization problem. Also since covariance matrices lie on 
non-linear Riemannian manifolds, we compare our former approach with a sparse 
linear approximation alternative that is suitable for equivalent vector spaces 
of covariance matrices. This is done by searching for the best projection of 
the query data on a dictionary using an Orthogonal Matching pursuit algorithm. 

We show the applicability of our video descriptor in two different application
domains - namely human action recognition and gesture recognition using one 
shot learning. Our experiments provide promising insights in large scale video 
analysis.
\end{abstract}

}
\maketitle

\IEEEdisplaynotcompsoctitleabstractindextext
\IEEEpeerreviewmaketitle

\section{Introduction}
\IEEEPARstart{E}{vent} recognition in unconstrained scenarios~\cite{REALACT,YOUTUBE,TRECVIDMED10,WEBDATA} 
has gained a lot of research focus in recent years with the phenomenal increase 
in affordable video content across the Internet. Most recognition algorithms 
rely on three important phases: extraction of discriminative low-level video 
features~\cite{STIP,SSTF,MBH}, finding a robust intermediate representation
~\cite{BOVW,MMI} of these features and finally, performing efficient  
classification. 

Feature extraction is unarguably very crucial for event recognition as 
introduction of noise at the earliest stage of the recognition process can 
result in undesirable performance in the final classification. Research in 
action or event recognition has addressed this problem in different 
ways. Early efforts include~\cite{STIP,SSTF} where the authors introduce 
special detectors capable of capturing salient change in pixel intensity 
or gradients in a space-time video volume and later describing these special 
points or regions using statistics obtained from neighboring pixels. Direct 
extension of interest point based approaches from images such as 
3D-SIFT~\cite{3DSIFT}(a space time adaptation of the SIFT~\cite{SIFT} 
descriptor), HOG3D~\cite{HOG3D}(a Spatio-Temporal Descriptor based on 3D 
Gradients derived from the principles of the HOG~\cite{HOG} descriptor for 
Human detection), Hessian STIP~\cite{HesSTIP} (a Hessian extension of the 
SURF~\cite{SURF} key-point detector to incorporate temporal 
discriminativity); are some of the proposed alternatives. Recently, Weng and 
colleagues introduced motion boundary histograms~\cite{MBH} that exploits the 
motion information available from dense trajectories.

\begin{figure*}[!ht]
  \begin{center}
  \includegraphics[width=0.99\textwidth]{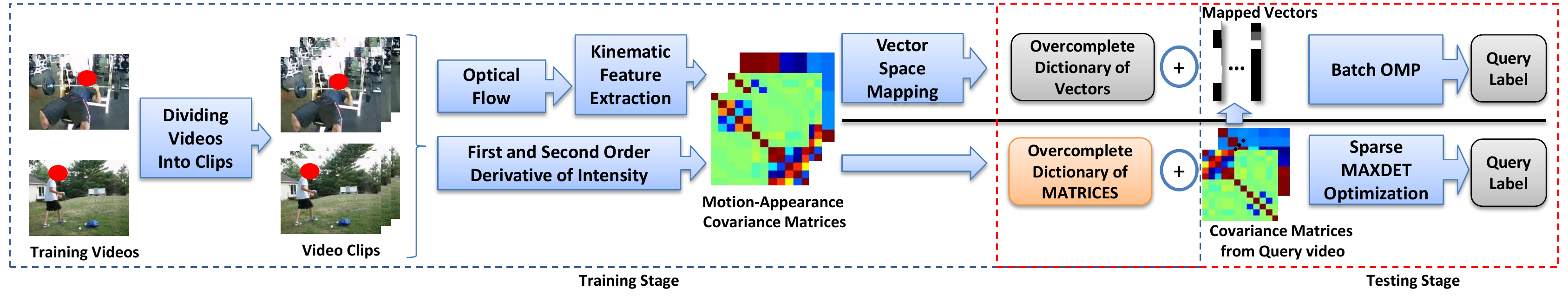}
  \caption{\footnotesize{\textbf{An Overview of our approach:} We begin with 
  dividing training videos into multiple non-overlapping clips. Each clip 
  is represented by a single covariance matrix computed from appearance 
  and motion features as explained in Sections ~\ref{sect:featcomp} and 
  ~\ref{sec:covcomp}. A dictionary is created by stacking up the covariance
  matrices. Given, a test covariance matrix, its corresponding label is 
  determined by solving a matrix determinant maximization problem as shown
  in Section~\ref{sect:tsc}. The final label for a video is obtained by 
  aggregating the class labels predicted for individual clips. 
  \label{fig:appr}}}
  \end{center}
  \vspace{-0.3in}
\end{figure*}

These interest point based approaches are incorporated into a traditional bag 
of video words framework~\cite{BOVW} to obtain an intermediate representation 
of a video that can further be used in a supervised~\cite{LIBSVM} or
un-supervised classification~\cite{DANN} algorithm for recognition purposes. 
While these approaches have been proved to be successful in context of event
recognition, since they rely on highly localized statistics over a small  
spatio-temporal neighborhood~\cite{SSTF,MBH} e.g. $50\times50\times20$
relative to the whole video, different physical motions within this small 
aperture, are indistinguishable. 

\begin{figure}[!ht]
  \begin{center}
  \subfigure[]{\label{fig:STIP}
  \includegraphics[width=0.48\textwidth]{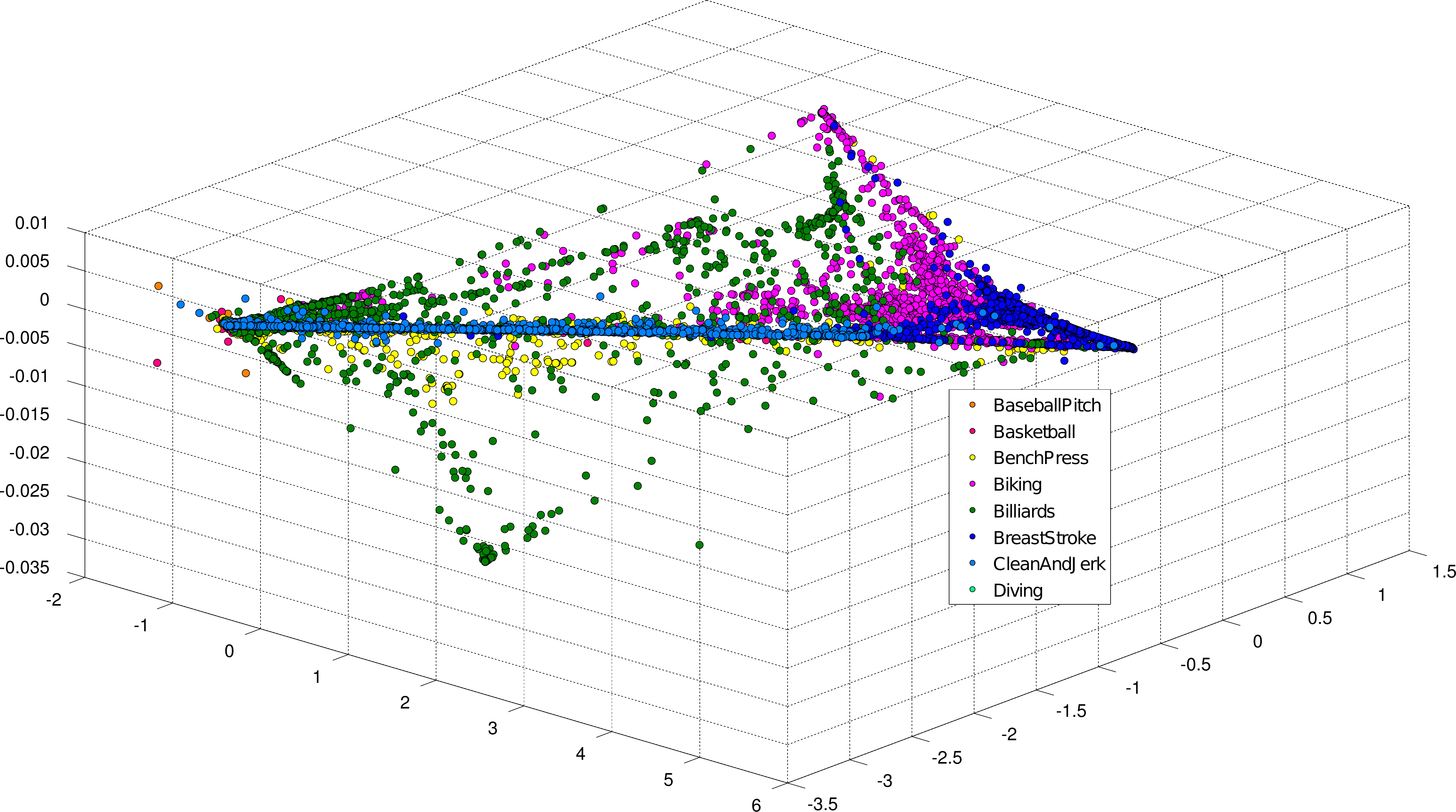}}
  \subfigure[]{\label{fig:COV}
  \includegraphics[width=0.45\textwidth]{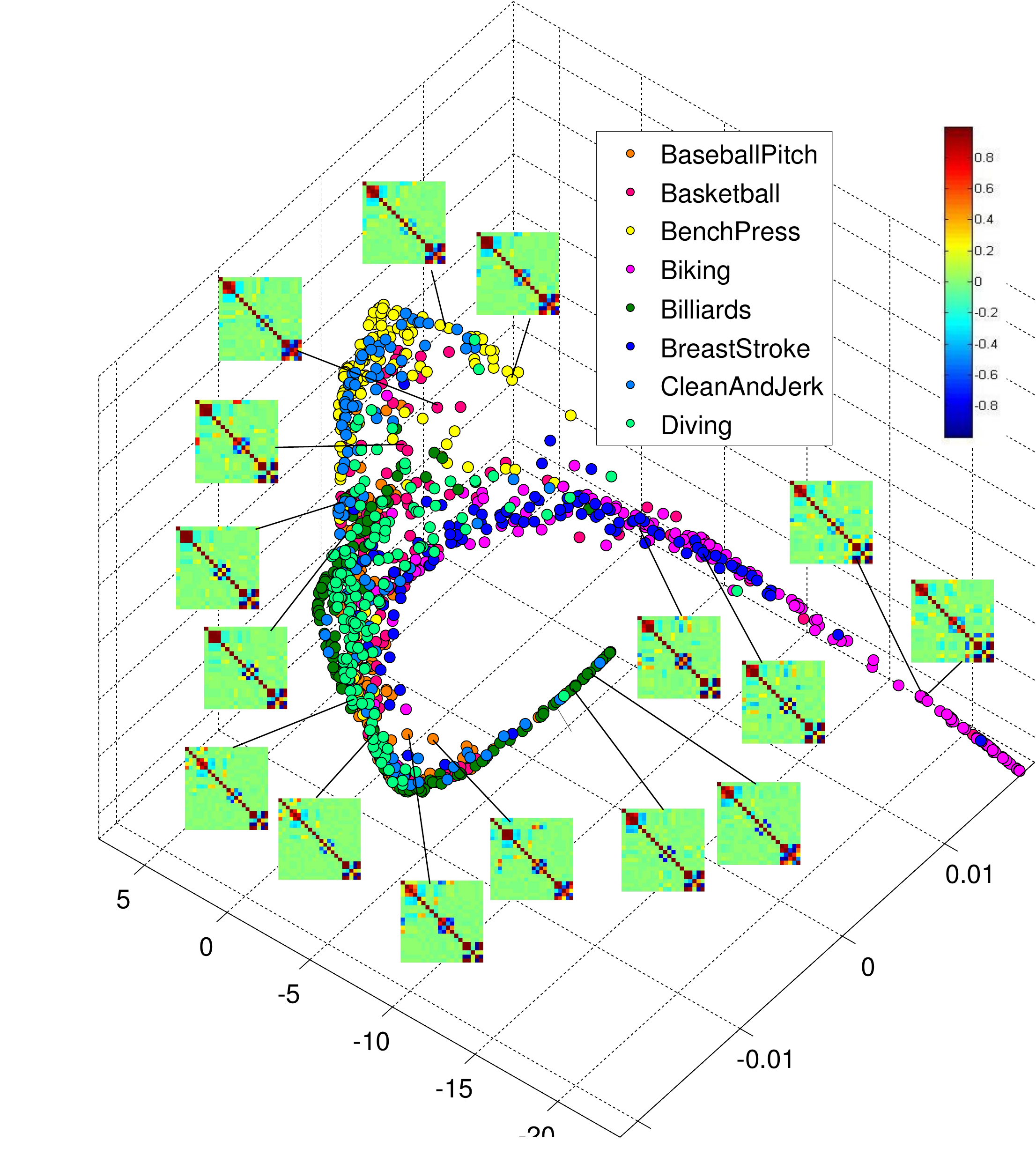}}
  \caption{\footnotesize{\textbf{Proposed descriptor vs HOG-HOF~\cite{STIP}:}
  Each circle represents a lower-dimensional manifestation of descriptors 
  from video samples in UCF50 human actions dataset. For legibility, only $8$ 
  classes are shown here using different color codes.~\subref{fig:STIP} 
  HOG-HOF~\cite{STIP} descriptors, originally $162$ dimensional, are reduced 
  to $3$-dimesional for visualization.~\subref{fig:COV} Proposed descriptor, 
  first vectorized ($190$ dimensional) then mapped to a $3$ dimesional 
  space. Sample covariance matrices are shown as insets to some circular 
  dots. Note how HOG-HOF based descriptors clutter the $3$-D space more 
  as compared to the proposed descriptors, which on the other hand, show 
  relatively clear cluster boundaries. Also, fewer descriptors are shown 
  in Fig.~\subref{fig:COV} which are clip based, as compared to 
  Fig.~\ref{fig:STIP} which are interest point based. See 
  Fig.~\ref{fig:appmot} for further detailed interpretation.
  \label{fig:combined}}}
  \end{center}
  \vspace{-0.2in}
\end{figure}

Also, while describing the statistics of these small neighborhoods, often the 
temporal and the spatial information are treated independently. For e.g. the 
HOG-HOF descriptor used in~\cite{STIP} is generated by concatenating two 
independent histograms : the HOG contributing to the appearance (spatial) and 
the HOF contributing to motion (temporal). Doing so, the joint statistics 
between appearance and motion is lost, particularly in case of human action 
and gesture recognition tasks, where such information can be very useful. 
For example, consider the example of ``pizza-tossing event'' from the 
UCF50~\footnote{http://vision.eecs.ucf.edu/data/UCF50.rar} unconstrained 
actions dataset. Here, a circular white object undergoes a vertical 
motion which is discriminative for this event class. Precisely, the correlation
between white object as captured by appearance features and its associated 
vertical motion captured basic and kinematic features is well explained 
in the covariance matrix than a concatenated 1-D histogram of the individual 
features. It is also important to note that contextual information available 
in the form of color, gradients etc., is often discriminative for certain 
action categories. Descriptors that are extensively gradient based such as HOG 
or HOF, need to be augmented with additional histograms such as color 
histograms to capture this discriminative information. 

In view of the above, we propose a novel descriptor for video event 
recognition which has the following properties: (1) Our descriptor is a concise
representation of a temporal window/clip of subsequent frames from a video 
rather than localized spatio-temporal patches, for this reason, we do not 
need any specialized detectors as required by~\cite{STIP,SSTF,HesSTIP}, (2) It 
is based on an effective fusion of motion features such as optical flow and 
their derivatives, vorticity, divergence etc., and appearance feature such as 
first and second order derivatives of pixel intensities, which are 
complementary to each other. This enables the descriptor to be extended to 
capture other complementary information available in videos e.g. audio, camera 
motion, very easily, (3) As the descriptor is based on joint distribution of 
samples from a set of contiguous frames without any spatial subsampling, it is 
implicitly robust to noise resulting due to slight changes in illumination, 
orientation etc. (4) It is capable of capturing the correlation between 
appearance with respect to motion and vice-versa in contrast to concatenated 
1-D histograms as proposed in~\cite{STIP,SSTF,3DSIFT,HOG3D}, also, since our 
final descriptor is based on the eigenvectors of the covariance matrix, they 
automatically transform our random vector of samples into statistically 
uncorrelated random variables, and (5) Finally being compact, fewer descriptors 
are required to represent a video compared to local descriptors and they need 
not be quantized. Fig.~\ref{fig:combined} provides an insight on the 
discriminative capability of both the HOG-HOF based descriptors and the 
proposed covariance matrix based descriptors. 

It is the semi-global, compact nature of our descriptor (since it is computed 
at clip level), that facilitates us to eliminate vector quantization based 
representation stage which is required in conventional bag-of-visual-words  
based frameworks, predominantly used in case of local descriptors
~\cite{STIP,SSTF,MBH}. Intuitively, we are interested to explore how 
contributions of constituent clips can be leveraged to categorize an entire 
video. In typical sparse representation based classification schemes~\cite{SPARSEFACE,SPARSEOBJ}, 
this issue is well-addressed. This motivates us to explore two sparse 
representation based techniques to perform event recognition using these 
covariance matrices as atoms of an over-complete dictionary. In the first one, 
we map the covariance matrices to an equivalent vector space using concepts 
from Riemannian manifold before building the dictionary. The classification is 
performed using a modified implementation of Orthogonal Matching 
Pursuit~\cite{OMP} which is specifically optimized for sparse-coding of large sets
of signals over the same dictionary. We compare this approach with a tensor 
sparse coding framework~\cite{MAXDET} formulated as a determinant maximization 
problem, which intrinsically maps these matrices to an exponential family. 
Although, our work is largely inspired by~\cite{RCOV} and ~\cite{MAXDET} in 
object recognition, to the best of our knowledge, ours is the first work that 
addresses event recognition using a sparse coding framework based on 
covariance of motion and appearance features.

The rest of this paper is organized as follows: Sect.~\ref{sect:relwork} 
discusses some of the related work in this direction. In the next section,
we provide the theoretical details of our approach including motion and appearance 
feature extraction, covariance computation followed by the sparse coding 
framework for classification. Next, we discuss two interesting applications and 
provide experimental details on how our descriptor and the classification methods 
can be applied to address these problems. Finally, Sect.~\ref{sect:conc} concludes 
the paper with future directions.

\section{Related Work}
\label{sect:relwork}
Covariance matrices as feature descriptors, have been used by computer vision 
researchers in the past in a wide variety of interesting areas such as: 
object detection~\cite{RCOV,COVLIC,COVHUMAN,COVLOGIT}, face recognition 
\cite{RCOVFR,MAXDET}, object tracking \cite{RCOVTR,RCOVTROCC}, etc. The authors
of ~\cite{RCOV} introduced the idea of capturing low-level appearance based 
features from an image region into a covariance matrix which they used in a 
sophisticated template matching scheme to perform object detection. Inspired 
by the encouraging results, a license plate recognition algorithm is proposed 
in~\cite{COVLIC} based on a three-layer, 28-input feed-forward back propagation 
neural network. The idea of object detection is further refined into human 
detection in still images~\cite{COVHUMAN} and videos~\cite{COVLOGIT}. In 
\cite{COVHUMAN}, Tuzel~\etal represented the space of d-dimensional nonsingular
covariance matrices extracted from training human patches, as connected
Riemannian manifold. A priori information about the geometry of 
manifold is integrated in a Logitboost algorithm to achieve impressive 
detection results on two challenging pedestrian datasets. This was later 
extended in~\cite{COVLOGIT} to perform detection of humans in videos, 
incorporating temporal information available from subsequent frames. 

The authors of~\cite{RCOVFR} used the idea of using region covariance matrices
as descriptors for human faces, where features were computed from responses of 
Gabor filters of $40$ different configurations. Later, Sivalingam~\etal 
proposed an algorithm~\cite{MAXDET} based on sparse coding of covariance 
matrices extracted from human faces, at their original space without 
performing any exponential mapping as proposed in previous approaches
~\cite{RCOV,COVLIC,COVHUMAN,COVLOGIT,RCOVFR}. In their approach, the authors 
formulated the sparse decomposition of positive definite matrices as convex 
optimization problems, which fall under the category of determinant 
maximization (MAXDET) problems. 

In a different vein, Porikli and Tuzel~\cite{RCOVTR} came up with another 
application of region covariance matrices in context of tracking detected
objects in a video. In their technique, the authors capture the spatial and 
statistical properties as well as their correlation of different features in 
a compact model (covariance matrix). Finally, a model update scheme is 
proposed using the Lie group structure of the positive definite matrices which 
effectively adapts to the undergoing object deformations and appearance changes. 
Recently, Li and Sun~\cite{RCOVTROCC} extended the tracking framework proposed
in~\cite{RCOVTR}, by representing an object as a third order tensor, further
generalizing the covariance matrix, which in turn has better capability to
capture the intrinsic structure of the image data. This tensor is further 
flattened and transformed to a reduced dimension on which the covariance 
matrix is computed. In order to adapt to the appearance changes of the object 
across time, the authors present an efficient, incremental model update 
mechanism.

\begin{figure*}[!ht]
  \begin{center}
  \subfigure[]{\label{fig:appearence}
  \includegraphics[width=0.283\textwidth]{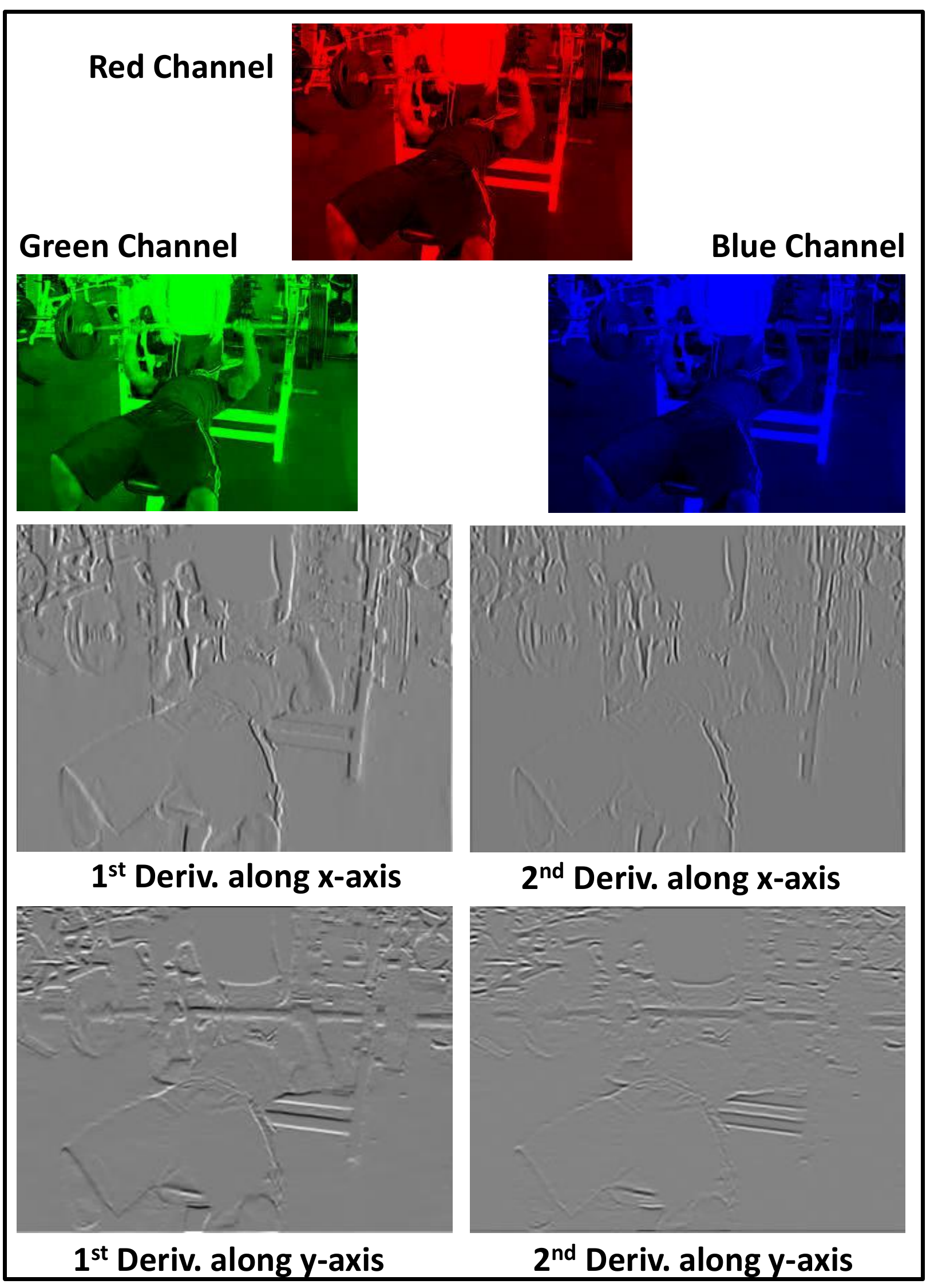}}
  \subfigure[]{\label{fig:motion}
  \includegraphics[width=0.682\textwidth]{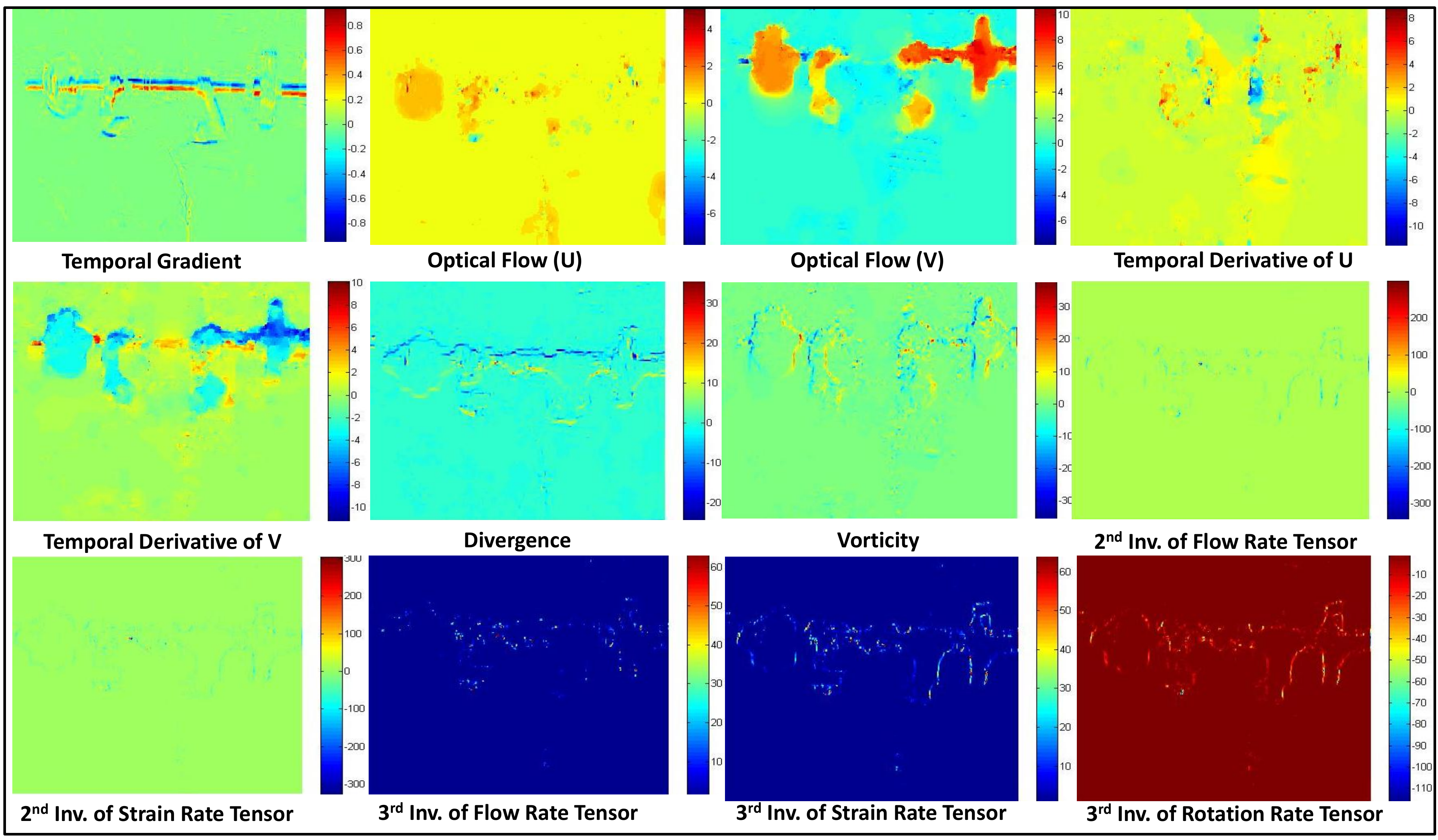}}
  \caption{\footnotesize{\textbf{Low-level feature extraction from video 
clips:} \subref{fig:appearence} Appearance features,and~\subref{fig:motion}
Motion features (basic and kinematic). The kinematic features are derived from 
optical flow and capture interesting aspect of motion with respect to a spatial 
neighborhood.}}
  \end{center}
  \vspace{-0.2in}
\end{figure*}

That said, in context of human action and gesture recognition, the 
exploitation of covariance matrices as feature is relatively inchoate. Some 
earlier advances are discussed here in this particular direction in order to 
set the pertinence of this work to the interested reader. Along these lines,
the authors of~\cite{COVFIRE} proposed a methodology for detection of 
fire in videos, using covariance of features extracted from intensities, 
spatial and temporal information obtained from flame regions. A linear SVM was 
used to classify between a non-flame and a flame region in a video. 
Researchers~\cite{COVSIL,COVSP} have also attempted to classify simple human 
actions~\cite{KTH} using descriptors based on covariance matrices. In 
contrast, our work addresses a more diverse and complex problem. To summarize, 
we make the following contributions in this work: (1) We propose a novel 
descriptor for video analysis which captures spatial and temporal variations 
coherently, (2) Our descriptor is flexible to be used for different application
domains (unconstrained action recognition, gesture recognition etc.), and (3) We 
extensively evaluate two different classification strategies based on concepts 
from sparse representation that can be used in the recognition pipeline 
independently.

\section{Our Approach}
\label{sect:approach}
In order to make the paper self contained, we briefly describe the theoretical 
details of all the phases involved in our action recognition computation pipeline,
beginning with the feature extraction step. Fig.~\ref{fig:appr} provides a 
schematic description of our approach showing the steps involved in training 
phase (dashed blue box) and the testing phase (dashed red box).

\subsection{Feature Computation}
\label{sect:featcomp}
Since our primary focus is on action recognition in unconstrained scenarios, we 
attempt to exploit features from both appearance and motion modalities which 
provide vital cues about the nature of the action. Also since this paper 
attempts to study how the appearance and motion change with respect to each 
other, it is important to extract features that are discriminative within a 
modality. Given a video, we split it into an ensemble of non-overlapping clips 
of $N$ frames. For every pixel in each frame, we extract the normalized 
intensities in each channel, first and second order derivatives along the $x$ 
and $y$ axes. Thus every pixel at $(x,y,t)$ can be expressed in the following 
vector form with $\mathbf{f_i}$, $\mathbf{f_g}$ denoting the color and 
the gray-scale intensity gradient components along the horizontal and vertical 
axes respectively, as:

\begin{eqnarray}
  \mathbf{f_i} &=& \left[ R \quad G \quad B \right]^T,\nonumber\\
  \mathbf{f_g} &=& \left[ \frac{\partial I}{\partial x} \quad 
                                    \frac{\partial I}{\partial y} \quad 
                                    \frac{\partial^2 I}{\partial x^2} \quad 
                                    \frac{\partial^2 I}{\partial y^2} \right]^T,
  \label{eqn:fa}
\end{eqnarray}
where $R,G,B$ are the red, green, blue intensity channels and $I$ being the 
gray scale equivalent  of a particular frame.
 
As motion in a video can be characterized using simple temporal gradient (frame
difference), horizontal ($u$) and vertical ($v$) components of optical flow vector, 
we use the following as our basic motion features: 
\begin{equation}
  \mathbf{f_m} = \left[ \frac{\partial I}{\partial t}  \quad u \quad v \quad \frac{\partial u}{\partial t}
                 \quad \frac{\partial v}{\partial t} \right]^T,
  \label{eqn:fm}  
\end{equation}
where $\frac{\partial}{\partial t}$ represents the finite differential operator 
along the temporal axis. In  addition to these basic flow features, we extract 
high-level motion features \cite{KINEMAT} derived from concepts of fluid 
dynamics, since these are observed to provide a holistic notion of pixel-level 
motion within a certain spatial neighborhood. For e.g. features such as 
divergence $\nabla$ and vorticity $\Gamma$ quantify the amount of local 
expansion occurring within flow elements and the tendency of flow elements to 
``spin'', respectively. Thus
\begin{eqnarray}
  \nabla&=&\frac{\partial u}{\partial x}+\frac{\partial v}{\partial y},\nonumber\\
  \Gamma&=&\frac{\partial v}{\partial x}-\frac{\partial u}{\partial y}.
  \label{eqn:dv}
\end{eqnarray}

Local geometric structures present in flow fields can be well captured by tensors 
of optical flow gradients~\cite{KINEMAT}, which is mathematically defined as:
\begin{equation}
  G = \begin{pmatrix}
       \frac{\partial u}{\partial x} & \frac{\partial u}{\partial y}\\
       \frac{\partial v}{\partial x} & \frac{\partial v}{\partial y}
       \end{pmatrix}.
  \label{eqn:FGT}
\end{equation}
With this intuition, we compute the principal invariants of the gradient tensor
of optical flow. These invariants are scalar quantities and they remain 
unchanged under any transformation of the original co-ordinate system. We
determine the second, $\tau_2(G)$ and third $\tau_3(G)$ invariants 
of $G$ as:
\begin{eqnarray}
  \tau_2(G) &=& \frac{1}{2}\left [tr(G)^2 +tr(G^2) \right],\nonumber\\
  \tau_3(G) &=& -det(G).
  \label{eqn:GTI}
\end{eqnarray}

Based on the flow gradient tensor, we determine the rate of strain, $S$ 
and rate of rotation, $R$ tensors which signify deviations from the 
rigid body motion, frequently seen in articulated human body movements. 
These are scalar quantities computed as :
\begin{eqnarray}
  S&=&\frac{1}{2}(G + G^T),\nonumber\\
  R&=&\frac{1}{2}(G - G^T).
  \label{eqn:RS}
\end{eqnarray}
Using the equations in(~\ref{eqn:GTI}), principle invariants can be computed 
for these tensors. The interested reader is requested to read~\cite{KINEMAT} 
for further insights on the selection of invariants. However, unlike the 
authors of~\cite{KINEMAT}, we do not compute the symmetric and asymmetric 
kinematic features as these assume human motion is centralized which 
is not valid for actions occurring in an unconstrained manner (typically seen
in YouTube videos). For the sake of legibility, we arrange the kinematic features 
computed from optical flow vectors in the following way,
\begin{equation}
  \mathbf{f}_k = \left[\nabla \quad \Gamma \quad \tau_2(G) \quad \tau_3(G) \quad
        \tau_2(S) \quad \tau_3(S) \quad \tau_3(R)\right]^T.
  \label{eqn:kmf}
\end{equation}

Finally we obtain the following representation for each pixel after 
concatenating all the above features to form a $19$ element vector as:
\begin{eqnarray}
  \mathbf{F} =  \left[\mathbf{f_i} \quad \mathbf{f_g} \quad \mathbf{f_m} \quad 
                      \mathbf{f}_k\right]^T.
  \label{eqn:F}
\end{eqnarray}
Figures~\ref{fig:appearence} and~\ref{fig:motion} visualize the appearance and 
motion features respectively for a sample frame from the UCF50 dataset, where
a person is exercising a ``bench-press''.

\begin{figure*}[!ht]
  \begin{center}
  \subfigure[]{\label{fig:apporig}
  \includegraphics[width=0.42\textwidth]{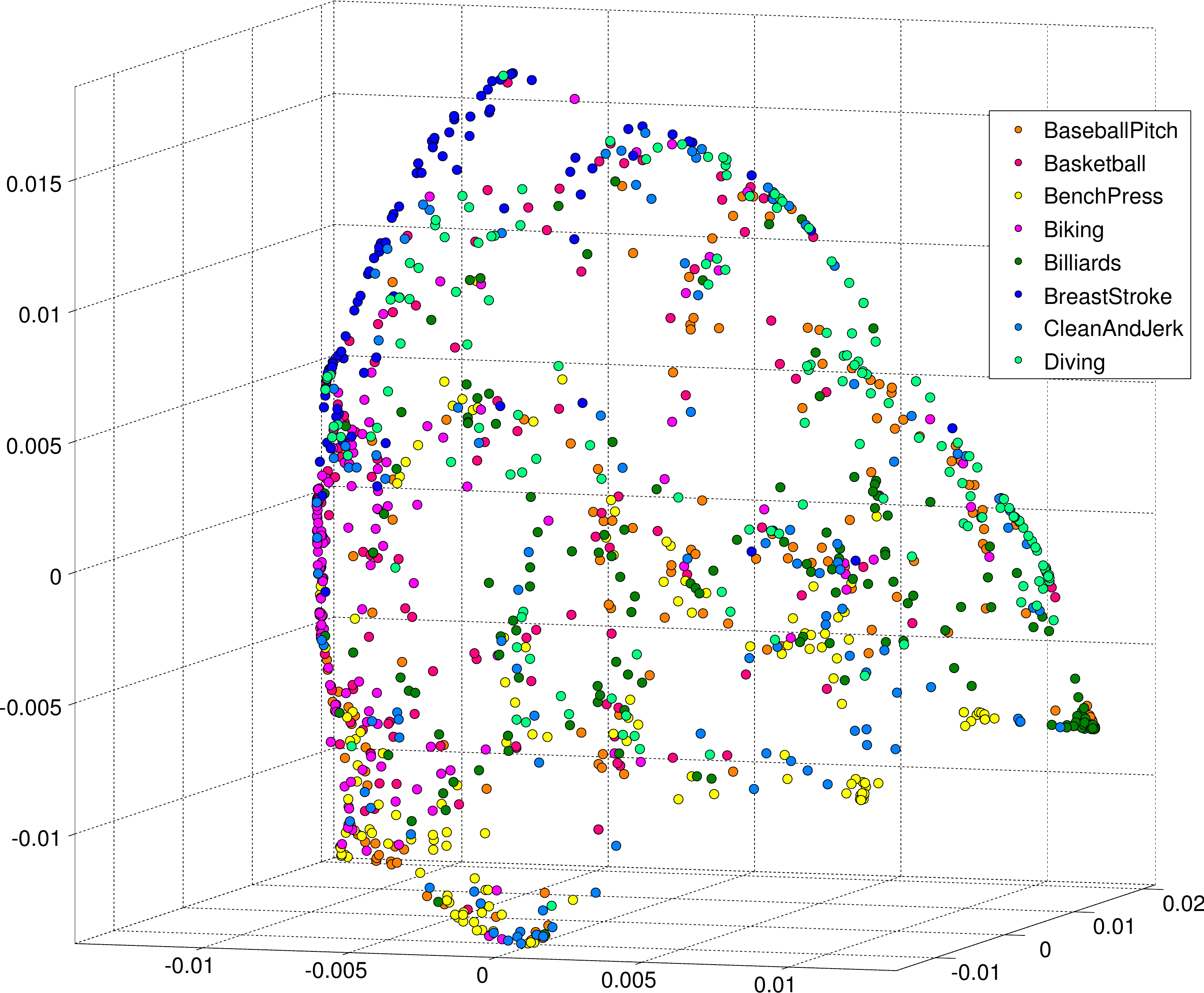}}
  \subfigure[]{\label{fig:applog}
  \includegraphics[width=0.42\textwidth]{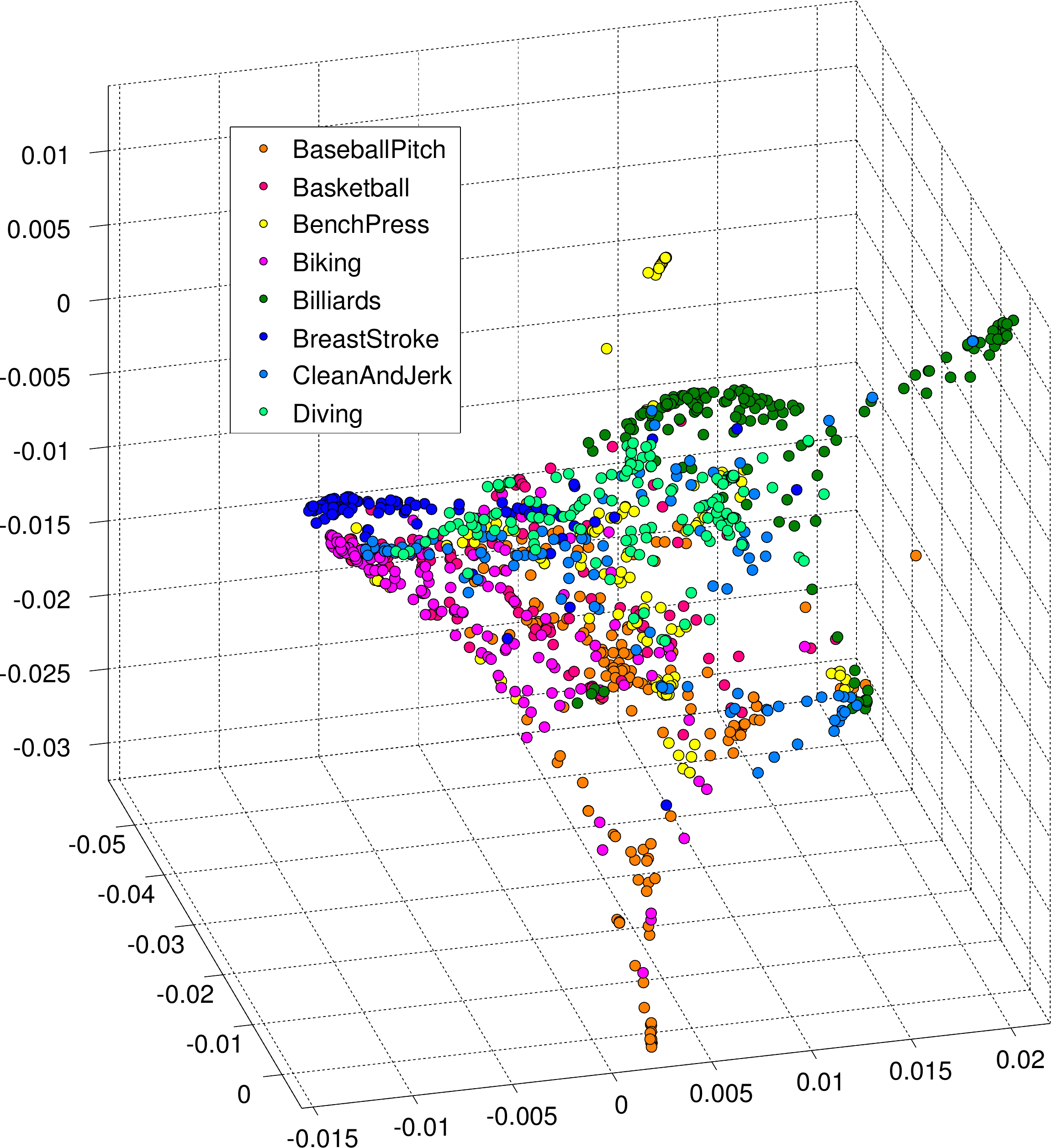}}
  \subfigure[]{\label{fig:motorig}
  \includegraphics[width=0.44\textwidth]{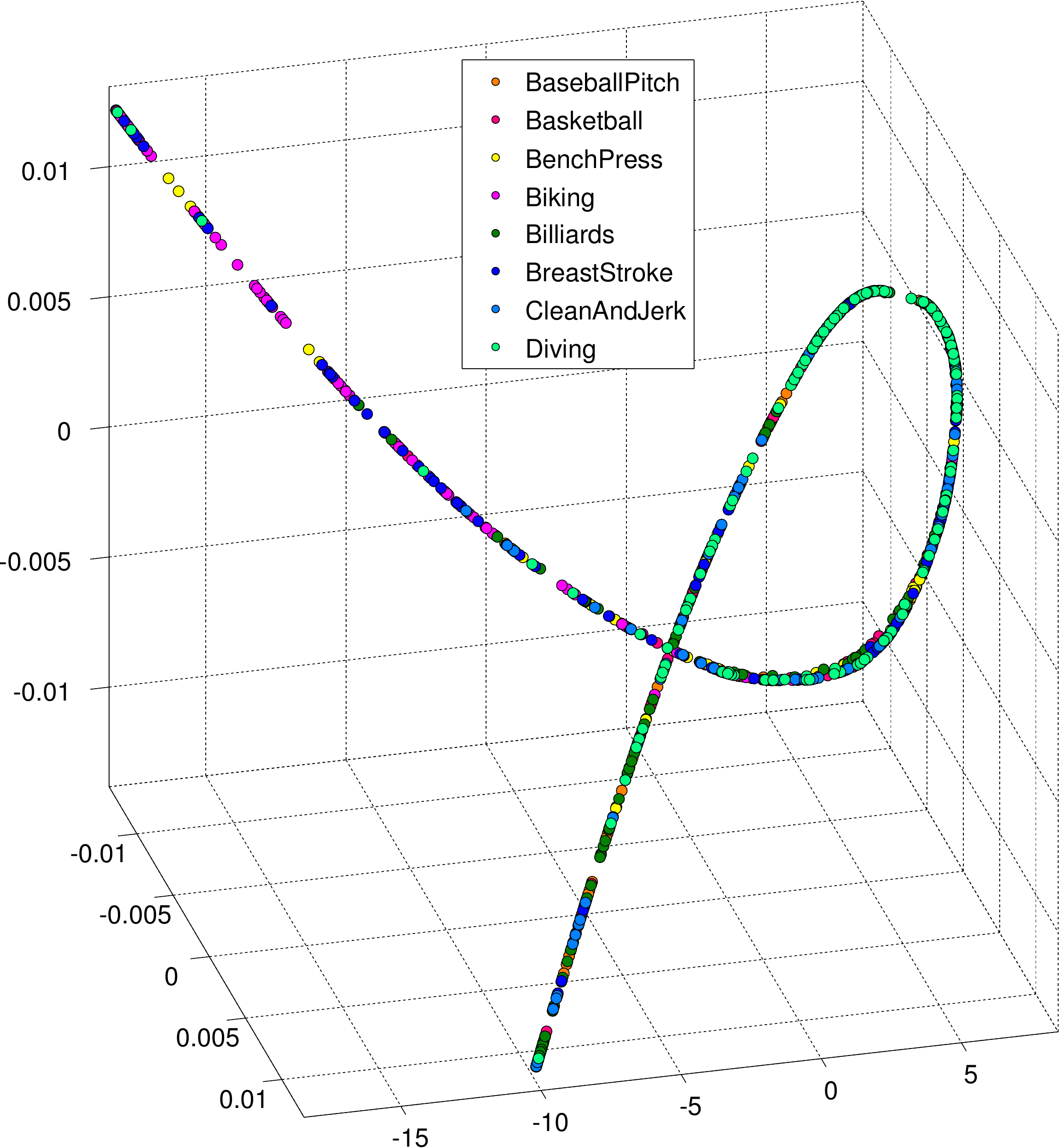}}
  \subfigure[]{\label{fig:motlog}
  \includegraphics[width=0.44\textwidth]{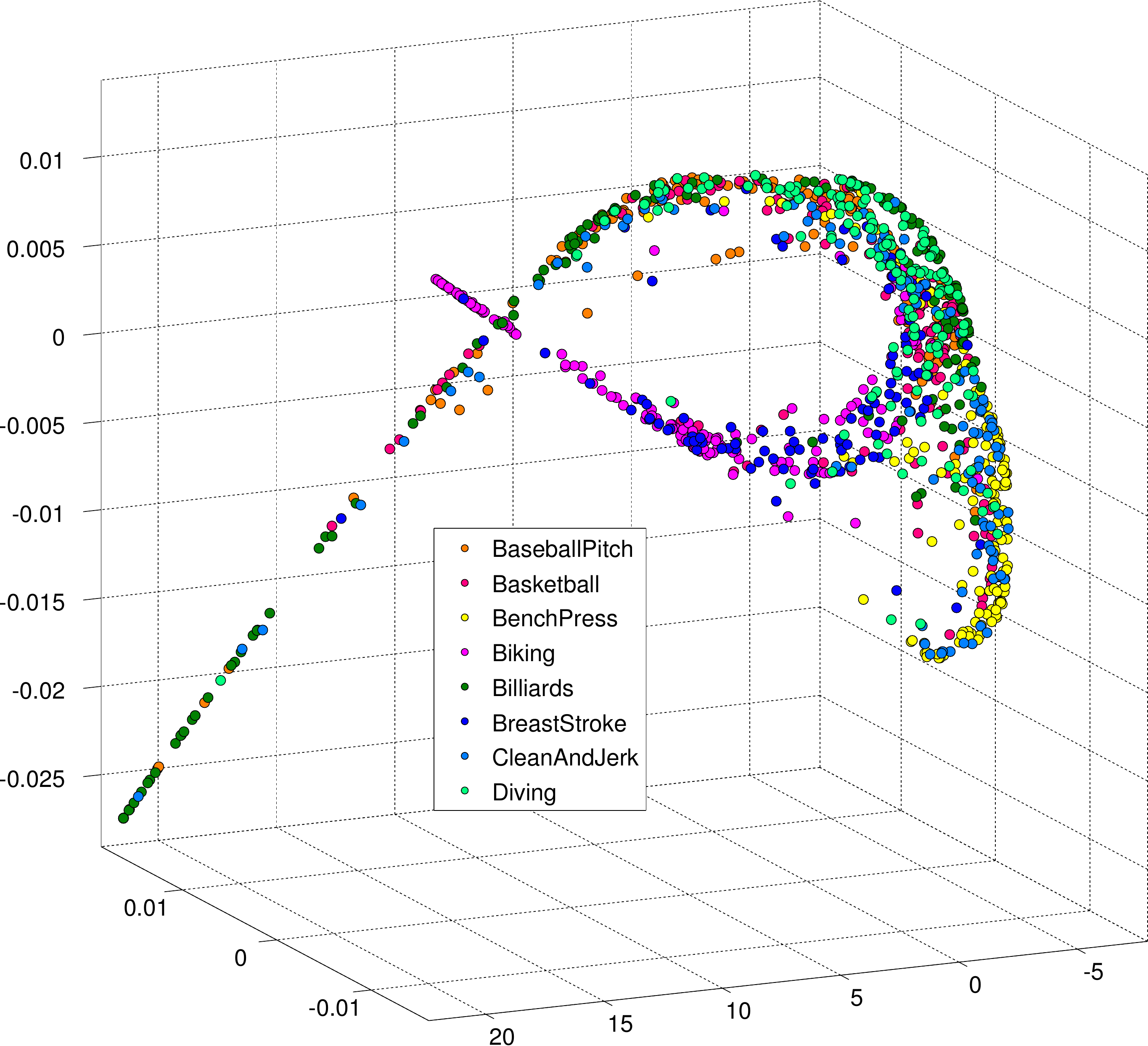}}
  \caption{\footnotesize{\textbf{Vector Space Mapping of $S_n^+$ covariance 
matrices}. The above two figures show covariance matrices (each matrix is a 
point) from video samples and their respective mapping in log-matrix space. In 
both the cases, representative video samples from $8$ arbitrary classes in 
UCF50 are chosen and their respective covariance matrices are determined. 
Different classes are colored differently. \subref{fig:apporig} shows original 
covariance matrices based on appearance features before mapping, and 
\subref{fig:applog} shows the same after mapping. \subref{fig:motorig} shows 
original covariance matrices based on motion features before mapping, and 
\subref{fig:motlog} shows the same after mapping. Note how some classes show 
more separability than others after the mapping. \label{fig:appmot}}}
  \end{center}
  \vspace{-0.2in}
\end{figure*}

\subsection{Covariance Computation}
\label{sec:covcomp}
Covariance based features introduced by Tuzel and colleagues for object recognition
~\cite{RCOV} have found application in various other related areas such as: 
face recognition~\cite{RCOVFR,MAXDET}, shape modeling~\cite{RCOVSM}, and 
object tracking~\cite{RCOVTR}. Based on an integral image formulation as 
proposed in~\cite{RCOV}, we can efficiently compute the covariance matrix 
for a video clip where each pixel is a sample. The covariance matrix in 
this context is therefore computed as : 
\begin{equation}
  C=\frac{1}{n-1}\sum_{i=1}^{n}(\mathbf{f^{(k)}_i}-\mu)(\mathbf{f^{(k)}_i}-\mu)^T ,
  \label{eqn:cov}
\end{equation}
where $\mathbf{f^{(k)}}$ is a single feature set and $\mu$ is its corresponding mean, 
$n$ being the number of samples (here pixels). Since the covariance matrix is 
symmetric, it contains $(d^2+d)/2$ ($d$ being the total types of features) 
unique entries forming the upper or lower triangular part of the matrix, that 
capture cross feature set variance.

\begin{figure*}[!ht]
  \begin{center}
  \includegraphics[width=0.98\textwidth]{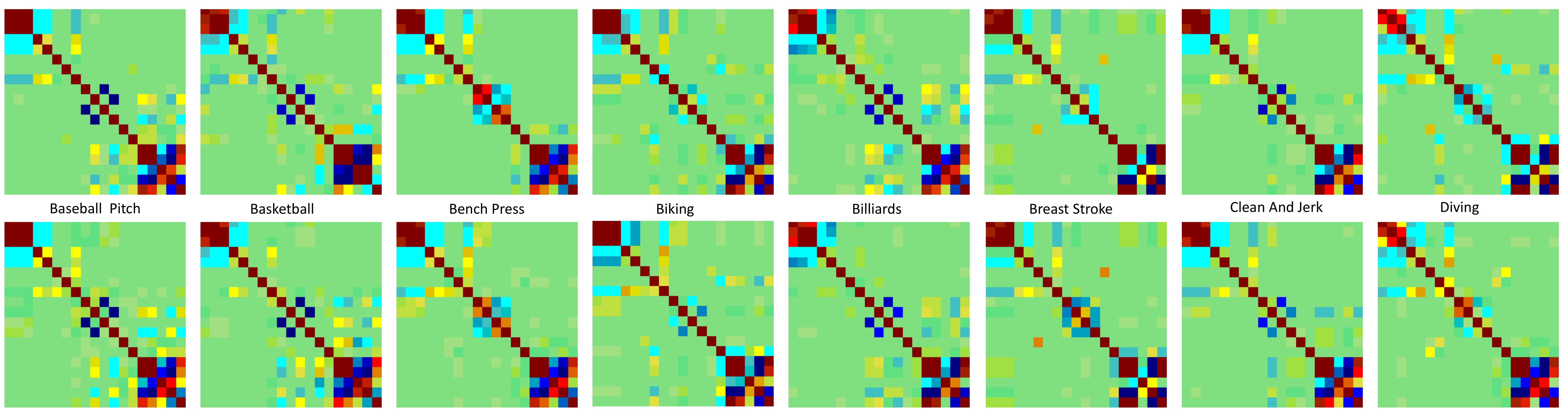}
  \caption{\footnotesize{\textbf{Normalized Covariance matrices from $8$ class 
of actions from UCF50:} Each column shows a different class, and each row 
is a sample covariance matrix constructed from clips belonging to one of the $8$ 
classes. We can notice the subtle differences between two samples of different 
classes and some structural similarity of elements of the same class. This 
aspect is more pronounced in Fig.~\ref{fig:combined}\label{fig:8classcovs}.}}
  \end{center}
  \vspace{-0.2in}
\end{figure*}

Covariance matrices have some interesting properties which naturally suits
our problem. Since these matrices do not have any notion of the temporal order 
in which samples are collected, they are computationally more favorable 
compared to trajectory based descriptors~\cite{MBH} that require explicit 
feature tracking. Secondly, covariance based descriptors provide a better way 
of analyzing relationship across feature sets compared to mere concatenation 
of histograms of different features~\cite{STIP}. Furthermore, the covariance
matrices provide more concise representation of the underlying feature 
distribution due to symmetry compared to long descriptors generated by 
methods proposed in~\cite{3DSIFT,SSTF} which need additional dimensionality 
reduction. We visualize descriptors computed from covariance matrices in 
figures~\ref{fig:appmot}, ~\ref{fig:8classcovs}.

Since, the covariance matrices are symmetric, either the upper or lower 
triangular elements can be used to form a vector $\in \mathbb{R}^l, 
l=(d^2+d)/2$ describing a clip. However, with that being said, vector addition 
and scalar multiplication $\in \mathbb{R}^l$, is not closed~\cite{COVMETRIC},
as the matrices conform to non-linear connected Riemannian manifolds of 
positive definite matrices ($S_n^+$). Hence, the descriptors obtained by 
direct vectorization as explained above, cannot be used as they are, for 
classification using regular machine learning approaches ($\mathbb{R}^2$). 
One possible approach to address this issue is to map these matrices to an 
equivalent vector space closed under addition or scalar multiplication, in 
order to facilitate classification tasks.

Fortunately, such an equivalent vector space for positive definite matrices 
exists, where these matrices can be mapped to the tangent space of the 
Riemannian manifolds~\cite{MATRIXLOG}. There are a couple of advantages of 
using this transformation, besides of the utility of being used in linear 
classification algorithms. The distance metric defined in this transformed 
subspace, is affine invariant and satisfies triangle inequality
~\cite{COVMETRIC}. Such transformation of a covariance matrix $C$ to its 
log $L$ can be performed using:
\begin{equation}
  L = \log(C) = R^T \hat{D} R, 
  \label{eqn:matlog} 
\end{equation}
where $R^T, R$ are rotation matrices obtained after singular value 
decomposition of $C$ and $\hat{D}$ is the diagonal matrix containing the log of 
eigenvalues. The mapping results in a symmetric matrix whose upper 
or lower triangular components form our final feature descriptor for a given 
video clip. 

Although these descriptors can be directly used in any vector 
quantization based bag-of-visual-words representation for classification tasks 
as used in ~\cite{STIP,MBH}, there is a major disadvantage. The matrix 
logarithm operation in Eqn.~(\ref{eqn:matlog}), due to its tangent space 
approximation of the original symmetric positive semi-definite space of 
covariance matrices, decimates structural information inherent to the 
matrices. Thus, further quantization performed in typical bag-of-visual-words 
based frameworks, can be detrimental towards the overall classification 
accuracy. We validate this empirically later in Sect.~\ref{sec:exset}. 
Therefore, we propose the use of sparse representation based techniques
for our classification problem, which eliminates further quantization of 
these descriptors, thereby leveraging on the existing available information. 

\subsection{Sparse coding of Matrix Log Descriptors}
\label{sec:scML}
Recently, sparse linear representation techniques have shown promising results 
in solving key computer vision problems including face recognition~\cite{SPARSEFACE},
object classification~\cite{SPARSEOBJ} and action recognition~\cite{COVSIL}.   
The basic objective of these approaches is to project the classification 
problem into a sparse linear approximation problem. Formally, given a set of $K$ 
training samples consisting of $k$ classes, $\mathbf{A}_1, \mathbf{A}_2, \ldots,
\mathbf{A}_K$ and a test sample $\mathbf{y}$, an over-complete dictionary $A$ is
constructed by stacking the training samples. Then the approximation problem: 
\begin{equation}
  \min ||\mathbf{x}||_1\quad s.t.\quad \mathbf{y} = A\mathbf{x}
  \label{eqn:sp}
\end{equation}
where $\mathbf{x}$ is a sparse vector of coefficients corresponding to each 
element in $A$, can be solved using linear programming techniques. For each 
coefficient in $\mathbf{x}$, the residuals : 
\begin{equation}
  r_i = ||\mathbf{y} - A\mathbf{x}_i ||_2 
  \label{eqn:res}
\end{equation}
are computed, where $\mathbf{x_i}$ is a zero vector with $i$th entry set to the 
$i$th coefficient in $\mathbf{x}$. The smallest residual identifies the true 
label of $\mathbf{y}$. 

Since, we have multiple descriptors per training sample, we modify the above
formulation to suit our problem in the following way: Given a set of $p$ 
clips from training videos, we construct our over-complete dictionary ($A$ 
in~(\ref{eqn:sp})) by stacking corresponding matrix log descriptors which are 
obtained after applying Eqn.~(\ref{eqn:matlog}). Thus, for a query video 
containing $m$ descriptors $\mathbf{y_1} \ldots \mathbf{y_m}$ from as many 
clips, our objective is to find how each of these clips can be efficiently 
approximated jointly using a linear combination of a subset of elements from 
$A$. Mathematically the problem can be stated as: 

\begin{equation}
  \mathbf{L} = 
    \left.
    \begin{array}{lll}
      \min ||\mathbf{y}_1 &-     &A\mathbf{x}||_2^2,\\
                          &\vdots&                  \\
      \min ||\mathbf{y}_m &-     &A\mathbf{x}||_2^2,
    \end{array}
    \right\} \text{s.t.} ||\mathbf{x}||_0\leq P,
  \label{eqn:spML}
\end{equation}
with $||\mathbf{x}||_0$ the $l_0$ pseudo-norm equal to the number of nonzero 
coefficients in $\mathbf{x}$, $P$ being an empirically determined threshold 
to control the degree of sparsity. Eqn.~(\ref{eqn:spML}) can be solved 
efficiently using batch version of the orthogonal matching pursuit~\cite{OMP}
\footnote{http://www.cs.technion.ac.il/$\sim$ronrubin/Software/ompbox10.zip}, 
which computes the residuals jointly $\forall \mathbf{y}_j$, by 
constraining the coefficients in $\mathbf{x}$ to be orthogonal projections 
of all clips in query sample $y$ on the dictionary $A$. Since each element in
$A$ is associated with a label indicating the class from which the clip is 
extracted, the solution to Eqn.~(\ref{eqn:spML}) yields $\mathbf{L} \in 
\mathbb{R}^m$, containing labels corresponding to each clip from the query 
video. The final label of the video can be obtained using a simple majority 
voting of the labels in $\mathbf{L}$. 

The technique discussed above can be viewed as a straight-forward solution to 
our problem. However, the above framework is only applicable to vector spaces.
Thus, although it retains more information as compared to vector quantization 
based methods in this context, it is unable to exploit the information 
available in the structure of the covariance matrices which conform to 
Riemannian geometry. This motivates us to explore further on the recent 
advances of Sivalingam and colleagues \cite{MAXDET} in sparse coding of 
covariance matrices which is discussed as follows.

\subsection{Tensor Sparse Coding of Covariance Matrices}
\label{sect:tsc}
Consider our query video consists of a single clip whose motion-appearance 
covariance matrix $Q$, constructed using Eqn.~\ref{eqn:cov}, can be expressed as 
a linear combination of covariance matrices forming an over-complete dictionary 
$D$:
\begin{equation}
  Q = x_1 D_1 + x_2 D_2+ \ldots + x_p D_p = \sum_{i=1}^{p}x_i D_i, 
  \label{eqn:comb}
\end{equation}
where $x_i$'s are coefficients of the elements $D_i$ from dictionary $D$ of 
covariance matrices of labeled training videos. As $Q$ belongs to the connected 
Riemannian manifold of symmetric positive definite matrices, the following 
constraint is implied:
\begin{equation}
  \hat{Q} \succeq 0, \Rightarrow  x_1 D_1+ x_2 D_2 + \ldots +x_p D_p \succeq 0, 
  \label{eqn:consympd}
\end{equation}
where $\hat{Q}$ is the closest approximation of $Q$, introduced to handle noise 
in real-world data. This closest approximation can be achieved by solving an
optimization problem. However, in order to perform this task, we first need to 
define a measure of proximity between our query matrix $Q$ and the approximated
solution $\hat{Q}$. Such a proximity measure is often measured in terms of 
penalty function called LogDET or Burg matrix Divergence~\cite{LOGDET} which is 
defined as:
\begin{equation}
  \Phi_{\nabla}(\hat{Q},Q) = tr(\hat{Q} Q ^{-1})-\log \det( \hat{Q} Q^{-1}) - d,
  \label{eqn:burg}
\end{equation}
Using Eqn.(\ref{eqn:comb}), the above equation can be further expanded as:
\begin{equation}
  \Phi_{\nabla}(Q,\hat{Q}) = tr(\sum_{i=1}^p x_i D_i {Q}^{-1})-\log \det(\sum_{i=1}^p x_i D_i {Q}^{-1}) - d,
  \label{eqn:exp} 
\end{equation}
Since, $\hat{D_i}= Q^{-1/2}D_i Q^{-1/2}$, we can substitute Eqn.(\ref{eqn:exp}) 
appropriately, achieving:  
\begin{eqnarray}
  \Phi_{\nabla}(Q,\hat{Q}) &=& tr(\sum_{i=1}^p x_i \hat{D}_i)-\log \det(\sum_{i=1}^p x_i\hat{D}_i) - d,\nonumber\\
                           &=& \sum_{i=1}^p x_i tr( \hat{D}_i)-\log \det(\sum_{i=1}^p x_i\hat{D}_i) - d,\nonumber\\
  \label{eqn:final}
\end{eqnarray}
where the $\log \det (.)$ function can be expressed as Burg Entropy of 
eigenvalues of a matrix $Z$ as $\log \det(Z) = \sum_i \log{\lambda_i}$. 
Therefore, our optimization problem can be formulated using the objective 
function in Eqn.(~\ref{eqn:final}) as: 
\begin{eqnarray}
  & \min_x &\sum_{i=1}^p x_i tr( \hat{D}_i)-\log \det(\sum_{i=1}^p x_i\hat{D}_i) + \delta||\mathbf{x}||_1 \nonumber\\
  & \text{subject to} & \mathbf{x}\geq 0  \nonumber, \quad \sum_{i=1}^p x_i\hat{D_i} \succeq 0, \text{and}, \quad \sum_{i=1}^p x_i\hat{D_i} \preceq I_n \nonumber\\
 \label{eqn:maxdet}
\end{eqnarray}
with, $\delta||\mathbf{x}||_1$ being a relaxation term that incorporates 
sparsity. The above problem can be mapped to a determinant maximization 
problem which can be efficiently solved by semi-definite programming 
techniques\footnote{http://cvxr.com/cvx/}. The optimization in Eqn.~
(\ref{eqn:maxdet}) can be performed separately for all $m$ clips in a 
video and the labels can be combined in the similar way as discussed in case 
of matrix log descriptors, leading to final label for a query sample. In the 
next sections, we provide our experimental details comparing the approaches 
presented here on two different application domains, finally discussing the 
results at the end of each sections.

\section{Experiments}
\label{sect:expr}
We organize this section into two parts that address two different problems
in video analysis encountered in practical scenarios. In the first one, we 
emphasize on action recognition in unconstrained case. The next part 
elucidates our observations on another important problem : one-shot 
recognition of human gestures. 

\subsection{Human Action Recognition}
\label{sect:per}
This is an extremely challenging problem, especially because videos depicting 
actions are captured in diverse settings. There are two newly introduced, 
challenging datasets (UCF50, HMDB51~\cite{HMDB}) containing videos that 
reflect such settings (multiple and natural subjects, background clutter, 
jittery camera motion, varying luminance). To systematically study the 
behavior of our proposed descriptor and the associated classification 
methods, we conduct preliminary experiments on a relatively simple, well 
recognized, human actions dataset~\cite{KTH} to validate our hypothesis and 
then proceed towards the unconstrained case.\\

\subsubsection{Datasets}
\noindent\textbf{KTH Human Actions:} This dataset~\cite{KTH} consists of $6$ 
classes namely: Boxing, Clapping, Jogging, Running, Walking, and Waving. The 
dataset is carefully constructed in a restricted environment -- clutter-free 
background, exaggerated articulation of body parts not seen in real-life,
mostly stable camera except for controlled zooming with single human actors. 
The videos in this dataset are in gray scale and not much cue is useful from 
background.\\

\noindent\textbf{UCF50:} The UCF50, human actions dataset consists of video 
clips that are sourced from YouTube videos (unedited) respectively. It consists
of over $6,500$ RGB video clips (unlike KTH) distributed over $50$  complex 
human actions such as horse-riding, trampoline jumping. baseball pitching, 
rowing etc. This dataset has some salient characteristics which makes 
recognition extremely challenging as they depict random camera motion, poor 
lighting conditions, huge foreground and background clutter, in addition to 
frequent variations in scale, appearance, and view points. To add to the above 
challenges, since most videos are shot by amateurs with poor cinematographic 
knowledge, often it is observed that the focus of attention deviates from the 
foreground.\\

\noindent\textbf{HMDB51:} The Human Motion DataBase~\cite{HMDB}, introduced in 
2011, has approximately $7,000$ clips distributed over $51$ human motion 
classes such as : brush hair, push ups, somersault etc. The videos have
approximately $640\times480$ spatial resolution, and are mostly sourced from 
TV shows and movies. The videos in the dataset are characterized by significant
background clutter, camera jitter and to some extent the other challenges 
observed in the UCF50 dataset.\\

\subsubsection{Experimental Setup}
\label{sec:exset}
We make some adjustments to the original covariance 
descriptor by eliminating appearance based features in Eqn.(\ref{eqn:F}) to 
perform evaluations on the KTH dataset, as not much contextual information 
is available in this case. Thus each pixel is represented by a $12$ dimensional
feature vector~(last $12$ features from $\mathbf{F}$ in~\ref{eqn:F}) resulting 
in a $(12^2+12)/2=78$ dimensional vector. Each video is divided into uniformly 
sampled non-overlapping clips of size $w\times h\times t$, $w,h$ being the 
original resolution of the video and $t$ is the temporal window. Throughout all 
experiments, we maintain $t=20$. Optical flow which forms the basis of our 
motion features, is computed using an efficient GPU implementation\cite{OFLOW}.

For all classification experiments we use a split-type cross-validation 
strategy suggested by the authors in~\cite{KTH}. We ensure that the actors that
appear in the validation set do not appear in the training set to construct
a dictionary for fair evaluation. Similar split strategy is employed for 
experiments on UCF50. For HMDB51 we follow the authors validation strategy that
has three independent splits. The average performance across all splits is 
recorded in Tables~\ref{tab:comp} and~\ref{tab:methods}.

In order to make fair comparison of our novel covariance based descriptor to 
a popular interest point based feature representation~\cite{STIP}, we use a 
traditional bag-of-visual-words framework for the latter. This forms our 
first baseline for all datasets (indicated as first row in Tab.
~\ref{tab:methods}). Next, we compare the proposed sparse representation based 
classification framework against three independent strategies,  using 
slightly different versions of our covariance descriptor. In the first, 
the covariance matrices are naively vectorized and the upper-triangular elements 
are used as clip-level descriptors. In the second, they are vectorized using the 
Eqn.~(\ref{eqn:matlog}) discussed in Sect.~\ref{sec:covcomp}. Each clip is 
used to train multi-class linear SVMs~\cite{LIBSVM} and for a query, labels 
corresponding to each clip are aggregated in a majority voting scheme (Sect.
~\ref{sec:scML}). In the next setting, we use a bag-of-visual-words framework 
for representing a video where the vocabulary is constructed by vector 
quantization of matrix log descriptors of covariance matrices. Experiments 
with different codebook sizes $64, 128, 256, 512, 1024, 2048$ are conducted.
Although the selection of codebook size is dataset specific, we observed 
recognition accuracies becoming asymptotic after relatively less codebook
sizes ($128$ for KTH, $512$ for both UCF50 and HMDB51). A histogram
intersection kernel based SVM is used as a final classifier. 

\begin{figure}[!ht]
  \begin{center}
  \includegraphics[width=0.48\textwidth]{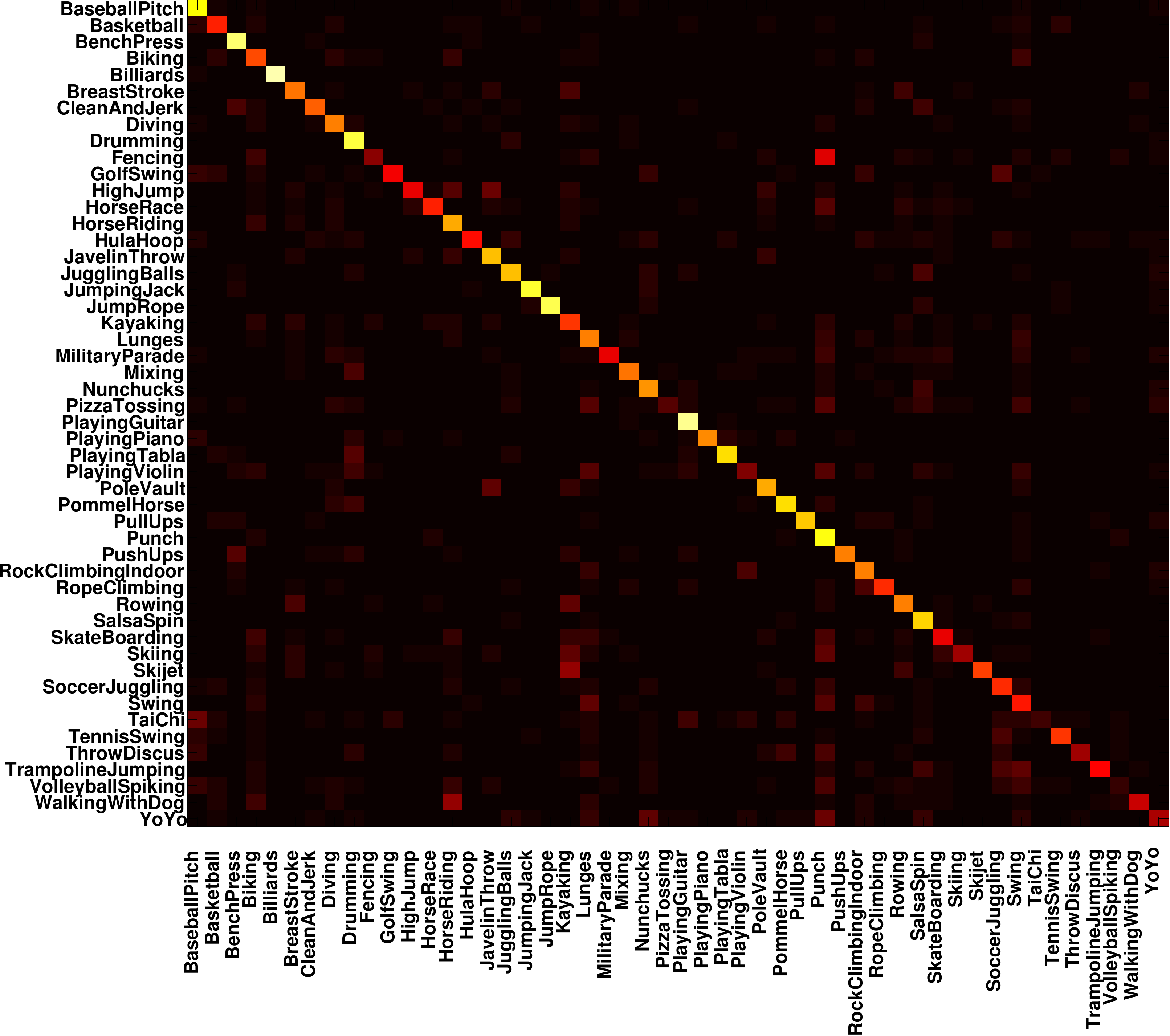}
  \caption{\footnotesize{\textbf{Confusion matrix} obtained after performing 
classification using the proposed classification technique on the UCF50.
\label{fig:UCF50}}}
  \end{center}
  \vspace{-0.2in}
\end{figure}

\begin{figure}[!ht]
  \begin{center}
  \includegraphics[width=0.48\textwidth]{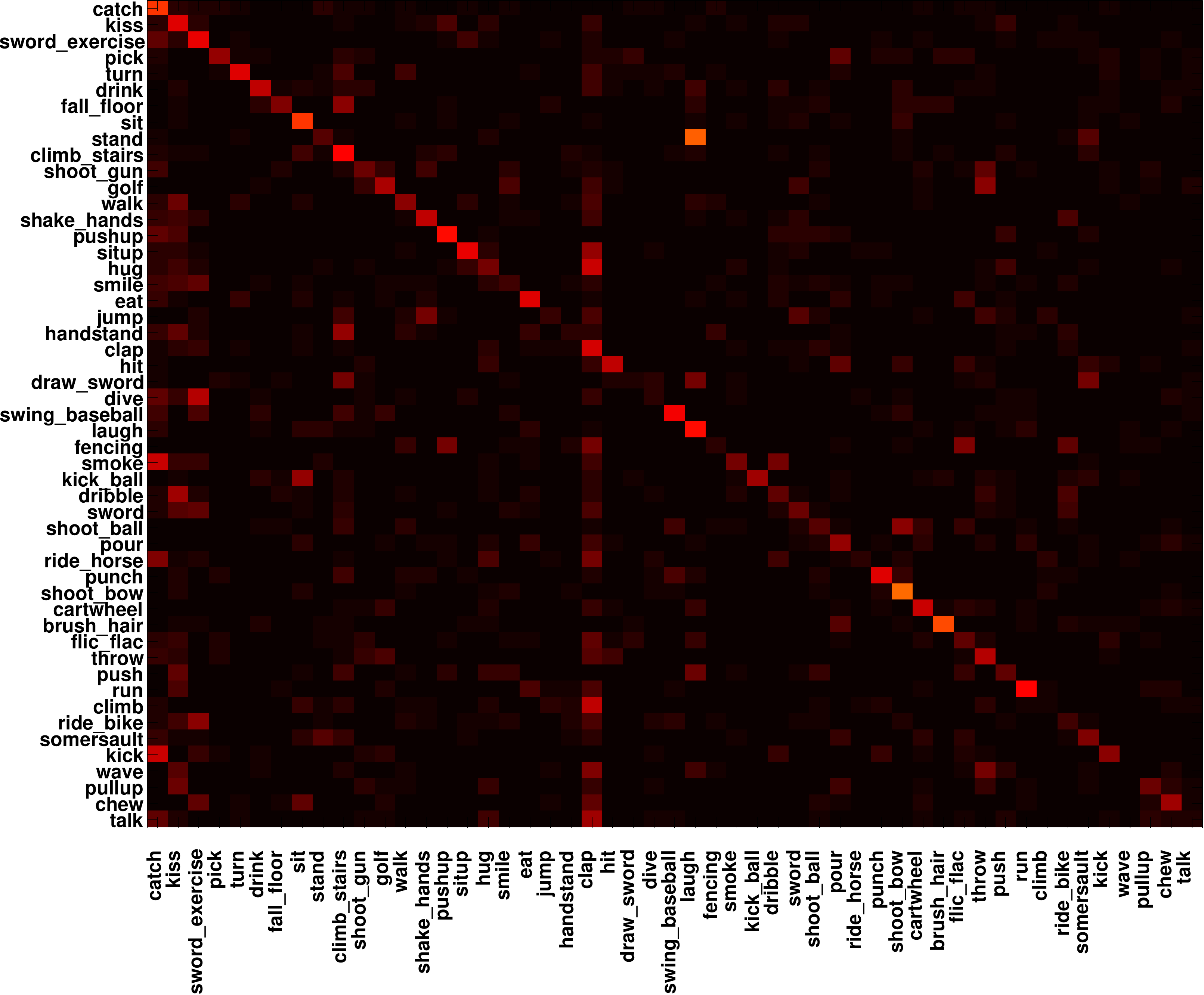}
  \caption{\footnotesize{\textbf{Confusion matrix} obtained after performing 
classification using the proposed classification technique on HMDB51 dataset.
\label{fig:HMDB51}}}
  \end{center}
  \vspace{-0.2in}
\end{figure}

\subsubsection{Results} 
In Tab.~\ref{tab:methods}, we present a comparative analysis of the various 
classification methods on these datasets. We compare our methods with the state
of the art performance obtained by other competitive approaches. Although our 
proposed method does not show significant improvement over the state of the art 
on the KTH dataset, we observe definite increase in performance over the two 
other challenging action recognition datasets. We also observe that there is a 
steady increase in performance across all datasets as we change our 
classification strategies that are more adapted to the matrix based 
descriptors which intuitively argues in favor of our original hypothesis. 
The reason can be attributed to vector quantization of the matrix based 
descriptors in the bag-of-visual-words representation (row $2-3$ of Tab.
~\ref{tab:methods}). Proper vectorization using the matrix log mapping 
(Eqn.~\ref{eqn:matlog}) increases the accuracy by $3-6\%$ (row $4$), which is 
further improved when sparse representation based classification is employed 
(row $5$). Finally, tensor sparse coding of covariance matrices (row $6$), 
achieves the best performance across all datasets. Note that the performance 
reflected in case of UCF50 and HMDB51 datasets are significantly high as 
compared to other approaches as a lot of contextual information is available 
from the RGB channels of the video. 

\begin{table}
  \begin{center} 
  \begin{tabular}{|l|l|c|c|c|}
     \hline
                    &                &\multicolumn{3}{c|}{\textbf{Datasets}}     \\
     \textbf{Method}& \textbf{Desc.} & \textbf{KTH}&\textbf{UCF50}& \textbf{HMDB51}\\    
     \hline
     BoVW        &HOG-HOF~\cite{STIP}& 92.0\%      &   48.0\%     &    20.2\%      \\
     BoVW           &COV             & 81.3\%      &   39.3\%     &    18.4\%      \\
     \hline
     SVM            &COV             & 82.4\%      &   40.4\%     &    18.6\%      \\
     SVM            &LCOV            & 86.2\%      &   47.4\%     &    21.03\%     \\
     \hline
     OMP            &LCOV            & 88.2\%      &   53.5\%     &    24.09\%     \\    
     TSC            &MAT             & 93.4\%      &   57.8\%     &    27.16\%     \\    
     \hline
  \end{tabular}
  \caption{\footnotesize{\textbf{Comparison with the state-of-the-art methods:} 
This table summarizes the performance of two of our proposed methods with 
different feature representations and classification strategies. First row 
shows the avg. recognition accuracies on a typical bag-of-visual-words 
framework on top of interest point based HOG-HOF descriptors~\cite{STIP}. 
In the next row, naively vectorized versions of the covariance matrices (COV) 
are used as descriptors in a similar BoVW framework. The next two rows 
reflect the same obtained after keeping linear SVM as classifiers using naive 
vectorization and matrix log descriptors (LCOV) as features. Finally the 
contributions of the proposed sparse representation based classification 
schemes -- Orthogonal Matching Pursuit (OMP) on LCOV and Tensor Sparse Coding 
(TSC) on the original covariance matrices (MAT) is highlighted in the two 
bottom rows, respectively.\label{tab:methods}}}
  \end{center}
\end{table}

\begin{figure*}[!ht]
  \begin{center}
  \includegraphics[width=0.98\textwidth]{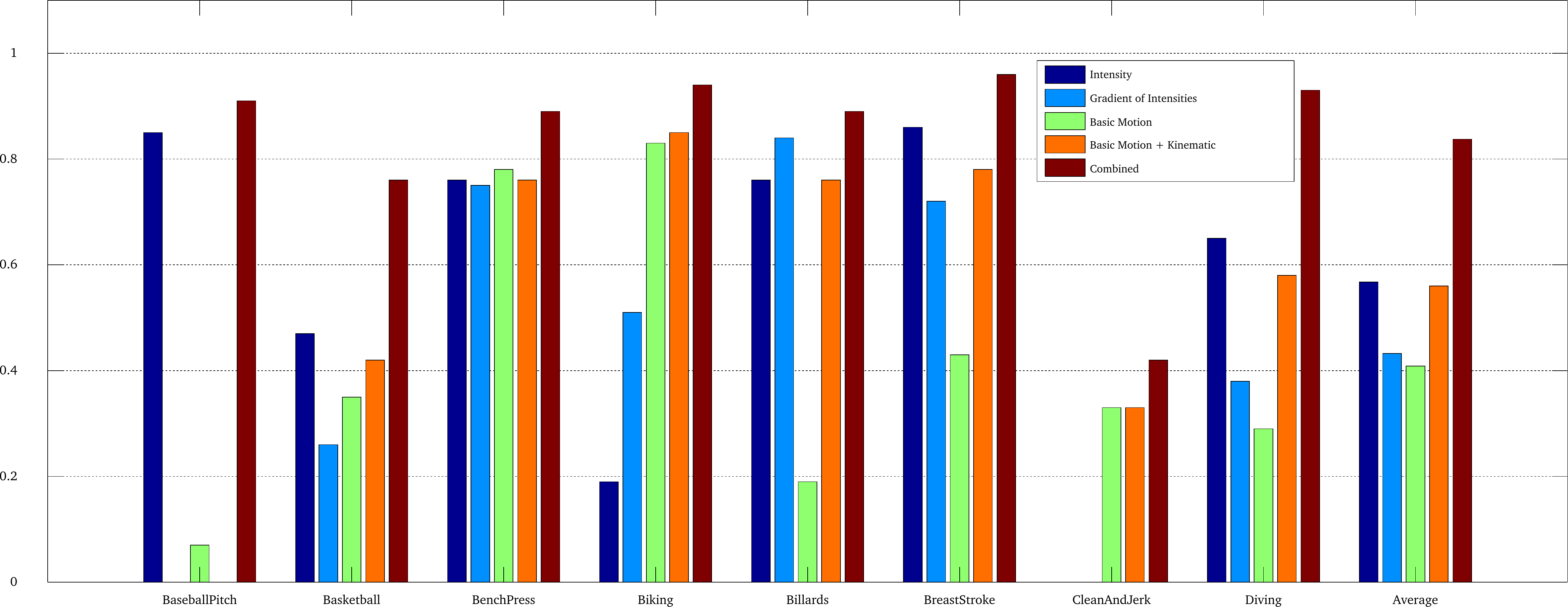}
  \caption{\footnotesize{\textbf{F-measures for $8$ classes from UCF50 dataset 
with different features}. The features experimented with are as follows : 
$\mathbf{F_i}$ (RGB Intensities), $\mathbf{F_g}$ (Intensity gradients), 
$\mathbf{F_m}$ (Basic Motion Features i.e. temporal derivatives, optical flow 
etc.), $\mathbf{F_k}$ (Kinematic Features), A combination of basic motion and 
kinematic features and finally all features are used together in the 
covariance descriptor.\label{fig:F1bar}}}
  \end{center}
  \vspace{-0.2in}
\end{figure*}

Given matrix log descriptors as feature, among linear SVM and OMP based 
classification, we observe OMP perform better than the former which shows that 
there is an inherent sparsity in the data which is favored by sparse 
representation based classification techniques. In Fig.~\ref{fig:UCF50} and 
Fig.~\ref{fig:HMDB51}, we present the confusion matrices obtained after 
classification using the tensor sparse coding which performs the best in case 
of both the datasets. In UCF50, the highest accuracies are obtained for 
classes that have discriminative motion (e.g. {\em Trampoline jumping} is 
characterized by vertical motion as opposed to other action categories). In 
case of action classes {\em Skijet}, {\em Kayaking} and {\em Rowing}, we 
observe high degrees confusion, as in all cases the low-level appearance 
components (water) in the covariance descriptors dominate over the motion 
components. A similar behavior is observed in case of two action classes in 
particular -- {\em Punch} and {\em Drumming} which show confusion with at 
least $5$ other classes which also occur mostly in indoor scenarios. In order 
to provide a profound insight, we analyze the individual low-level feature set 
contributions towards the recognition accuracies. 

\begin{figure}[!ht]
  \begin{center}
  \subfigure[]{\label{fig:roccaj}
  \includegraphics[width=0.48\textwidth]{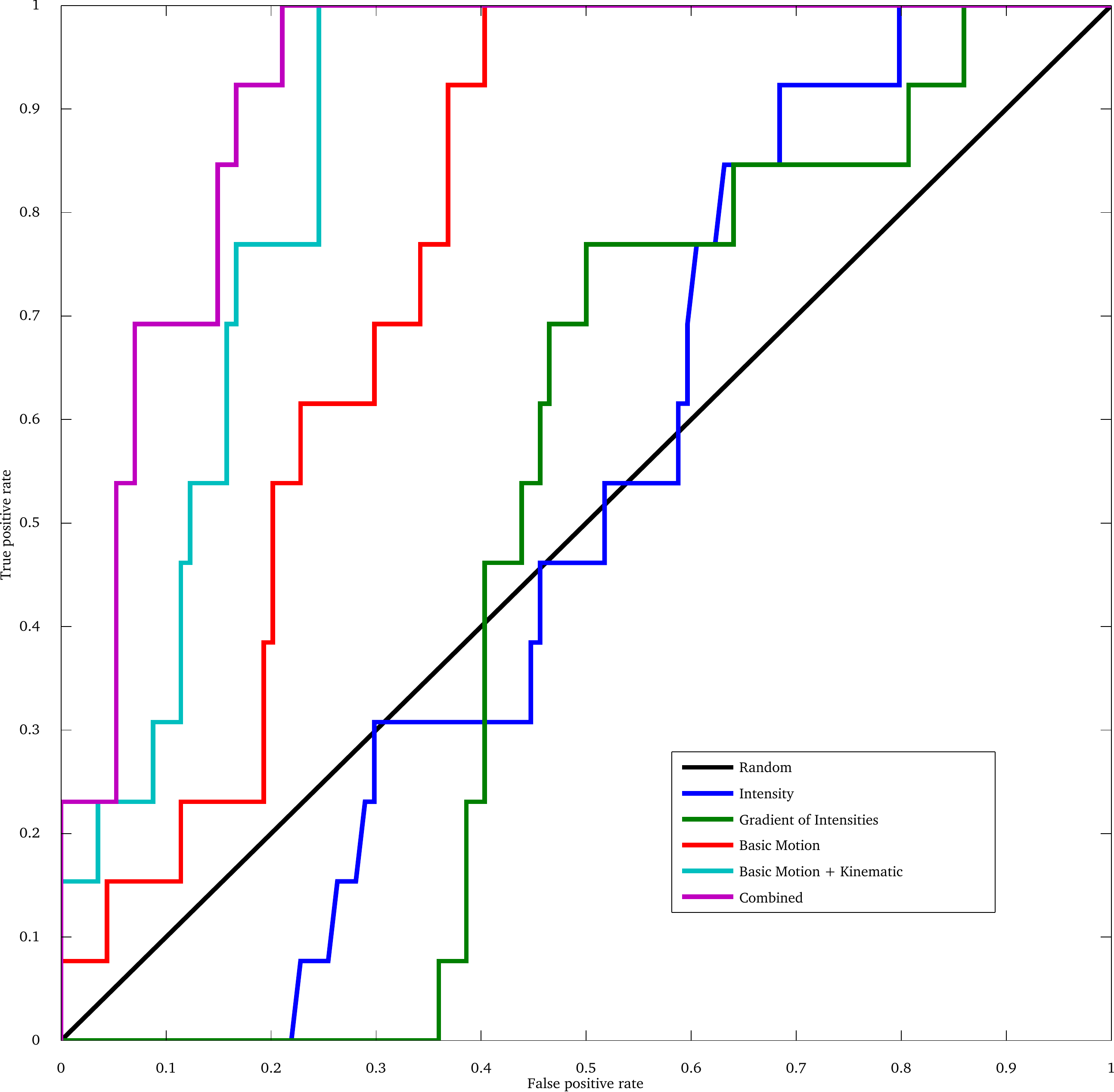}}
  \subfigure[]{\label{fig:rocbbp}
  \includegraphics[width=0.48\textwidth]{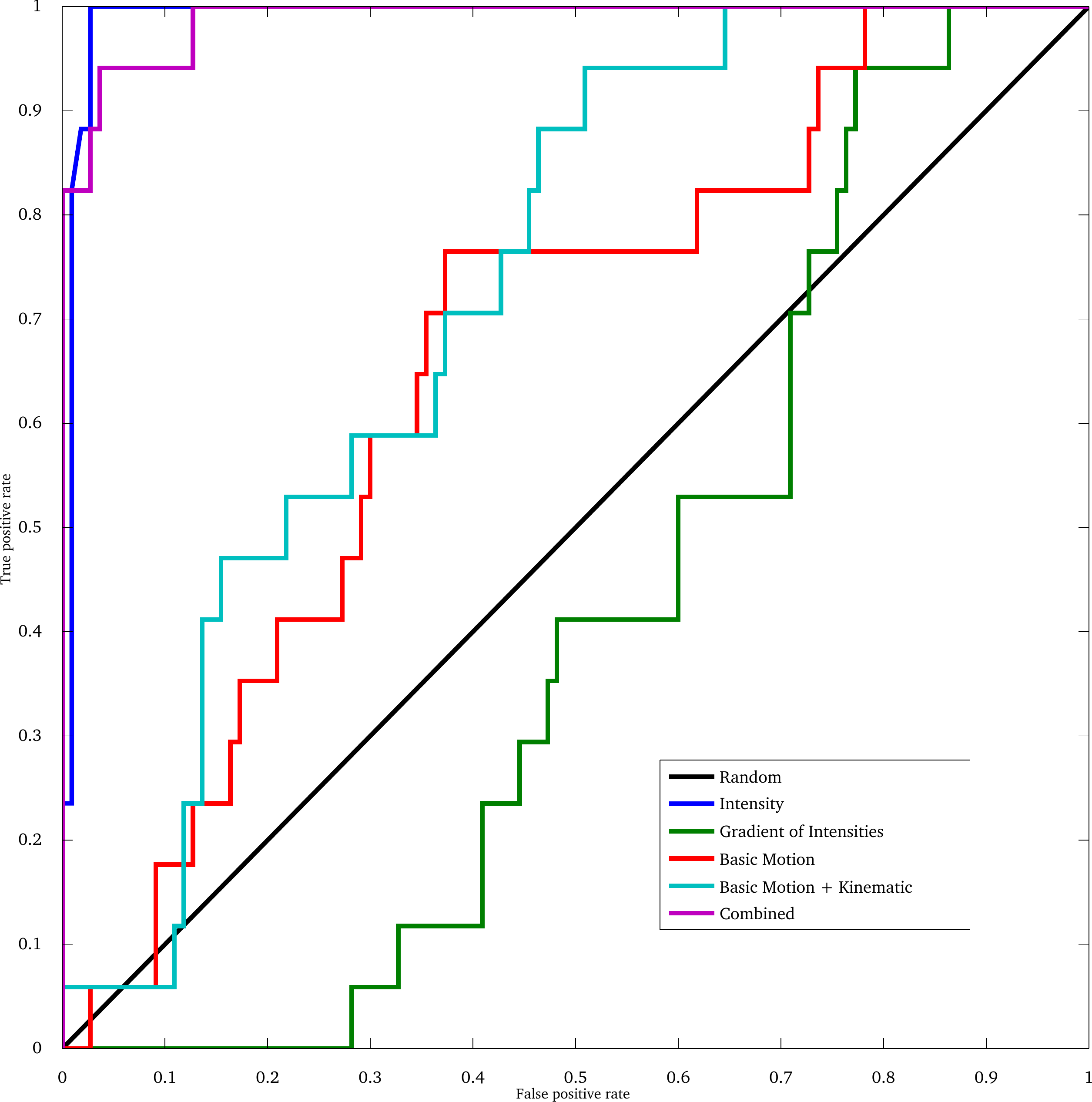}}
  \caption{\footnotesize{\textbf{Precision-Recall curves for detection} of 
\subref{fig:roccaj} {\em CleanAndJerk} and \subref{fig:rocbbp} {\em 
BaseballPitch} samples from UCF50. Of the $8$ classes analyzed in 
Fig.~\ref{fig:F1bar}, these are the two classes which have clear separation 
because of their distinctive motion features.\label{fig:roc}}}
  \end{center}
  \vspace{-0.2in}
\end{figure}

\noindent\textbf{Low-level Feature Set Contributions:} To investigate the 
contribution of different feature modalities towards the recognition 
performance, we computed $3$ different sets of covariance matrices 
for videos in UCF50. Firstly, descriptors computed using only appearance 
features (resulting in a $7\times7$ matrix). Next, we use only motion based 
features. Thus the covariance matrix in this case is  $12\times12$. Finally, 
both appearance and motion features are used together to compute the 
covariance matrices. We also evaluated how each classification strategy 
behaves with these different descriptors. For each of these descriptors, 
the classification framework was varied between a linear SVM (SVM/LCOV), 
Sparse OMP (OMP/LCOV), and finally the Tensor Sparse Coding (TSC) 
algorithm that uses MAXDET optimization. For the first two methods, the 
descriptors are of the following dimensions: $28$ (appearance), $78$ (motion) 
and $190$ (all).

We observed that the appearance features are less informative as compared to 
the motion features in videos where RGB information is available. However, all 
classification techniques get a boost in performance when both the features 
are used together, which shows how the proposed descriptor captures 
complementary information. Tensor Sparse Coding based classification performs 
better than other two methods. Tab.~\ref{tab:comp} summarizes the results of 
the experiments involving the contribution of different feature modalities and 
methods. The different columns in the table show the feature modalities used 
for computing the covariance matrices (AF = Appearance Features, MF = Motion 
Features, AMF = Appearance and Motion Features). 

The individual feature contribution towards the overall classification 
performance, is further experimented with finer granularity. Fig.
\ref{fig:F1bar} indicates F-measures derived from precision and recall for
 $8$ different classes of unconstrained actions from UCF50 dataset. It is 
interesting to notice two distinct trends from this experiment: RGB 
intensities contribute the most towards the discriminativity of the covariance 
descriptor for {\em Baseballpitch} while {\em CleanAndJerk} is best described 
by motion features. This can be explained by the sudden vertical motion captured 
by the basic motion and kinematic features in {\em CleanAndJerk} samples, and 
the mostly greener texture of background captured by intensity features in 
{\em Baseball-pitch} samples. The Precision-Recall curves for detection of these 
classes are provided in Fig.~\ref{fig:roc}, emphasize the contribution of the 
features in further finer granularity. The following section provides a brief 
discussion on the algorithmic complexities involved in the various steps of 
the entire recognition pipeline.

\begin{table}
  \begin{center} 
  \begin{tabular}{|l|l|c|c|c|}
     \hline
                     &                & \multicolumn{3}{c|}{\textbf{Experiments on UCF50}}\\
     \textbf{Method} & \textbf{Desc.} & \textbf{AF}  & \textbf{MF} & \textbf{AMF}\\    
     \hline
     SVM             & LCOV           & 31.4\%       &   43.4\%    &    47.4\%     \\
     LC/OMP          & LCOV           & 34.2\%       &   42.5\%    &    51.5\%     \\    
     TSC             & MAT            & 34.5\%       &   46.8\%    &    53.8\%     \\    
     \hline
  \end{tabular}
  \caption{\footnotesize{\textbf{Contribution of feature sets and methods on 
UCF50:} First row shows Matrix Log descriptors (LCOV) from covariances with 
linear SVM classifier, second row - LCOV with sparse OMP classifier, and 
finally Tensor Sparse Coding (TSC) on Covariance Matrices (MAT). Columns show 
the feature modalities used for computing the covariance matrices (AF = 
Appearance Features, MF = Motion Features, AMF = Appearance and Motion 
Features combined).\label{tab:comp}}}
  \end{center}
\end{table}

\subsubsection{Complexity Analysis}
The entire computation pipeline can be summarized in three major steps, namely
low-level feature extraction, feature fusion using covariance matrices, followed
by classification. Off these, the feature extraction and covariance computation
step for each clip of a video can be done in parallel for any dataset. Among 
feature extraction, optical flow computation~\cite{OFLOW} is the most expensive 
step, which is based on a variational model. For a consecutive pair of frames, 
with a resolution of $512\times384$, a GPU implementation of the above algorithm, 
takes approximately $5$ seconds on a standard desktop computer hosting a 2.2Ghz 
CPU with 4GB of physical memory. Depending on the types of low-level features 
computed, the complexity of the covariance matrix computation is $O(WHdC)$ 
where $d$ is the total types of low-level features, $W$ and $H$ are the 
respective width and height of a typical clip, and $C$ being the total number of
frames per clip.

The complexity of classification using the Orthogonal Matching Pursuit\cite{OMP} 
scheme is optimized using an efficient batch implementation provided in\cite{BOMP}. 
Since this method involves precomputation of an in-memory dictionary of fixed 
number of elements ($T_D$), the overall complexity can be approximated as $O(T_D + 
K^2d + 3Kd + K^3)$, where $K$ is the target sparsity for sparse coding. For 
details, please refer \cite{BOMP}. Classification using MAXDET optimization, on 
the other hand, is relatively more expensive as it attempts to find a subset of 
dictionary atoms representing a query sample using a convex optimization. In 
closed form, this is $O(d^2L^2)$, $L$ being the number of dictionary atoms.  
Although, this technique is more reliable in terms of accuracy, it requires a 
larger computation overhead as the process needs to be repeated for every query 
sample. Assuming the number of samples are far larger than $L$ batch-OMP is 
observed to offer a respectable trade-off between accuracy and speed.

\subsection{One-shot Learning of Human Gestures}
\label{sect:OLHG}
In addition, to demonstrate the applicability of our video descriptor, we 
report our preliminary experimental results on different application domain:
human gesture recognition using a single training example.  

\begin{figure*}[!ht]
  \begin{center}
  \includegraphics[width=0.8\textwidth]{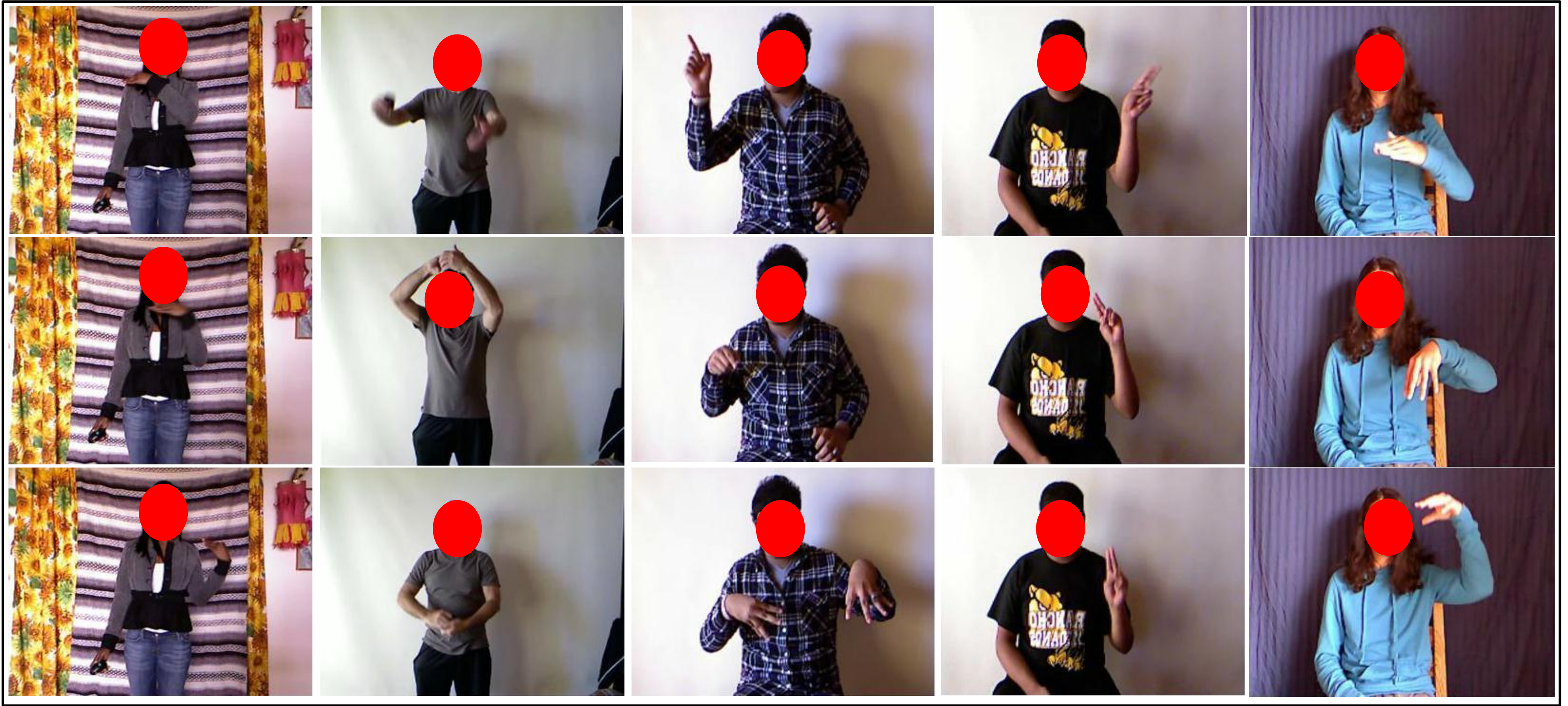}
  \caption{\footnotesize{Sample frames from representative batches from the 
CGD 2011 dataset.\label{fig:oneshot}}}
  \end{center}
  \vspace{-0.2in}
\end{figure*}

\subsubsection{ChaLearn Gesture Data (CGD) 2011} 
This dataset is compiled 
from human gestures sampled from different lexicons e.g. body language 
gestures (scratching head, crossing arms etc.), gesticulations performed to 
accompany speech, sign languages for deaf, signals (referee signals, diving 
signals, or marshaling signals to guide machinery or vehicle) and so on.
Within each lexicon category, there are approximately, $50$ video samples  
organized in different batches, captured using depth and RGB sensors provided 
by the Kinect~\footnote{http://en.wikipedia.org/wiki/Kinect} platform. Each 
video is recorded at $30$ Hz at a spatial resolution of $640\times480$. Each 
batch is further divided into training and testing splits and only a single 
example is provided per gesture class in the training set. The objective is 
to predict the labels for the testing splits for a given batch. 

Although the videos are recorded using a fixed camera under homogeneous lighting
and background conditions, with a single person performing all gestures within 
a batch, there are some interesting challenges in this dataset. These are 
listed as follows: (1) Only one labeled example of each unique gestures, (2)
Some gestures include subtle movement of body parts (numeric gestures), (3) 
Some part of the body may be occluded, and, (4) Same class of gesture can 
have varying temporal length across training and testing splits.

\begin{table}
  \begin{center} 
  \begin{tabular}{|l|c|c|c|c|}
  \hline
         & \multicolumn{4}{c|}{\textbf{Descriptor Performance(\%)}}\\
Batch ID & M      & MG      & MP   & All\\
  \hline
Devel01  & 66.7   &  66.7   & 88.3 &  83.3\\
Devel02  & 53.3   &  66.7   & 53.3 &  75.0\\
Devel03  & 28.6   &  42.9   & 21.4 &  28.6\\
Devel04  & 53.3   &  58.3   & 75.0 &  75.0\\
Devel05  & 92.8   &  100    & 92.8 &  100.0\\
Devel06  & 83.3   &  91.7   & 83.3 &  91.7\\
Devel07  & 61.5   &  76.9   & 61.5 &  84.6\\
Devel08  & 72.7   &  72.7   & 81.8 &  81.8\\
Devel09  & 69.2   &  61.5   & 69.2 &  69.2\\
Devel10  & 38.5   &  61.5   & 53.6 &  53.6\\
   \hline
Avg.     & 62.9   &  69.9   & 68.0 &  74.3\\    
   \hline
\end{tabular}
  \caption{\footnotesize{\textbf{Contribution of different low level features 
towards the one-shot gesture learning problem:} Each column shows a different 
set of low-level features used to compute the final descriptor. The order of 
low-level feature sets are as follows: Basic Motion (M), Basic Motion and 
Intensity gradients (MG), Subset of Basic motion and positional informations 
(MP), and finally all features combined. Refer to Section~\ref{sec:gcsetup} 
for more details\label{tab:desc}.}}
  \end{center}
\end{table}

\subsubsection{Experimental Setup}
\label{sec:gcsetup}
We obtain a subset of $10$ batches from the entire 
development set to perform our experiments. For a given batch, the position of 
the person performing the gesture remains constant, so we adjust our feature
vector in Eqn.(\ref{eqn:F}) to incorporate the positional information of the 
pixels $x,y,t$ in the final descriptor. Furthermore, since the intensities of 
the pixels remain constant throughout a given batch, the RGB values at the 
corresponding pixel locations could also be eliminated. Also, the higher order 
kinematic features such as $\tau_2(S)$, $\tau_3(S)$, and $\tau_3(R)$ can be 
removed as they do not provide any meaningful information in this context.
Thus each pixel is represented in terms of a $16$ dimensional feature vector, 
resulting in a $16\times16$ covariance matrix with only $136$ unique entries.
The upper triangular part of the log of this matrix forms our feature 
descriptor for a clip extracted from a video. In order to perform 
classification, we use a nearest neighbor based classifier with the same 
clip-level voting strategy as discussed in the earlier experiments. A regular 
SVM based classifier is not applicable to this problem as there is only one
training example from each gesture class.

Since depth information is available along with the RGB videos, we exploit it 
to remove noisy optical flow patterns generated by pixels in the background, 
mainly due to shadows. 

\begin{table}
  \begin{center} 
  \begin{tabular}{|l|c|c|c|c|}
  \hline
         & \multicolumn{4}{c|}{\textbf{Method Acc. Avg. (\%)}}\\
Batch ID & MBH/NN & STIP/NN & TPM  &  LCOV/NN\\
  \hline
Devel01  & 66.7   &  25.0   & 58.3 &  83.3\\
Devel02  & 33.4   &  8.3    & 25.0 &  75.0\\
Devel03  & 7.2    &  28.6   & 14.3 &  28.6\\
Devel04  & 33.4   &  16.7   & 58.4 &  75.0\\
Devel05  & 28.6   &  14.3   & 64.3 &  100.0\\
Devel06  & 50.0   &  16.7   & 25.0 &  91.7\\
Devel07  & 23.1   &  7.7    & 15.4 &  84.6\\
Devel08  & 36.4   &  9.09   & 9.1  &  81.8\\
Devel09  & 30.7   &  23.1   & 53.8 &  69.2\\
Devel10  & 15.4   &  15.4   & 23.1 &  53.6\\
   \hline
Avg.     & 32.4   &  16.4   &  34.7&  74.3\\    
   \hline
\end{tabular}
  \caption{\footnotesize{\textbf{Comparison with other features/methods}:
This table summarizes the performance of our descriptor in one-shot gesture 
recognition against other methods. The leftmost column contains the batches 
on which the methods are tested. The next two columns contain indicate 
the avg. accuracy obtained using two local feature based approaches: MBH
~\cite{MBH} and STIP~\cite{STIP} using a Nearest Neighbor (NN) classifier, the
next column uses template matching (TPM) based method and the last column 
records the performance of our descriptor (LCOV) when used with a nearest 
neighbor classifier.\label{tab:1shot}}}
  \end{center}
\end{table}

\subsubsection{Results} 
Similar to the previous experiments on action recognition in section
~\ref{sect:per}, we perform a detailed analysis, with more emphasis on the 
descriptor. To this end, we use different versions of the descriptor with only 
motion features (M:$9\times9$ covariance matrix), a combination of motion and 
intensity gradients (MG:$13\times13$ covariance  matrix), a combination of 
motion and positional information (MP:$12\times12$ covariance matrix) and 
finally all features combined ($16\times16$). The results are reported in Tab.
~\ref{tab:desc}. We observe that again motion in itself is not the strongest 
cue. However, when fused with appearance gradients and positional information,
 the overall performance of the descriptor increases by $11\%$, which is a 
significant improvement considering the nature of the problem. 

\begin{figure*}[!ht]
  \begin{center}
  \includegraphics[width=0.96\textwidth]{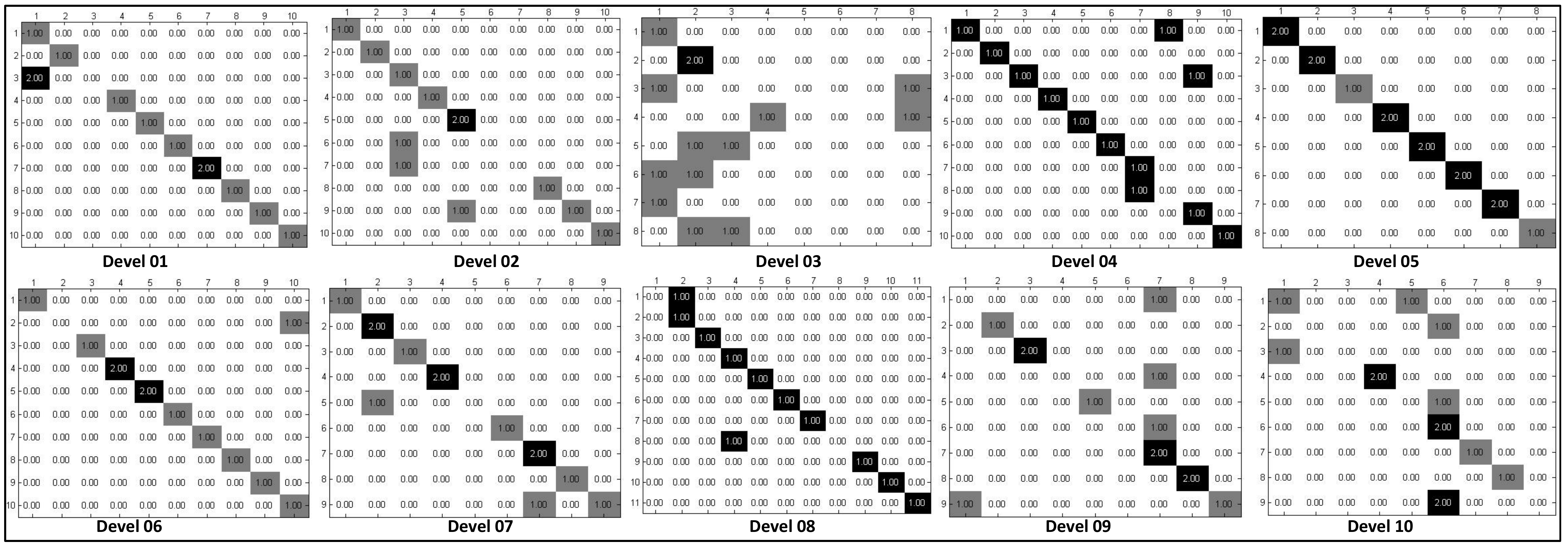}
  \caption{\footnotesize{Confusion matrices obtained after applying the 
proposed method on first $10$ of the development batches from the CGD 2011 
dataset. Note for certain batches (devel01--02, devel04--05, devel06--08), our 
method is able to predict gesture labels with respectable accuracies using just 
one training sample.\label{fig:gestconfmat}}}
  \end{center}
  \vspace{-0.3in}
\end{figure*}

In order to make a fair evaluation of our descriptor with the state-of-the-art 
descriptors from action recognition literature~\cite{STIP,MBH}, we keep the 
classifier constant (Nearest Neighbor). We also compared our approach with 
a simple template matching based recognition which is more appropriate for this 
type of problem. The average accuracies for each batch tested using all the 
compared methods are reported in Table~\ref{tab:1shot}. It is pleasing to note 
that our descriptor performs significantly better than all other methods which
gives us promising leads towards the applicability of this descriptor for this 
class of problems. Finally, in Fig.~\ref{fig:gestconfmat}, we show the 
respective confusion matrices obtained after applying the proposed method on 
first $10$ of the development batches from the CGD 2011 dataset.

\section{Conclusion \& Future Work}
\label{sect:conc}
We presented an end-to-end framework for event recognition in unconstrained scenarios. 
As part of this effort, we introduced a novel descriptor for general purpose video 
analysis that is an intermediate representation between local interest point based 
feature descriptors and global descriptors. We showed that how simple second order 
statistics from features integrated to form a covariance matrix can be used to perform
video analysis. We also proposed two sparse representation based classification 
approaches that can be applied to our descriptor. As part of future work, we intend 
to fuse more information in our proposed descriptor such as audio and would like
to explore on optimizing the MAXDET approximation problem which is currently a 
computationally expensive operation in our recognition framework.

\bibliographystyle{abbrv}
\bibliography{TPAMI2012mocov-v3}
\end{document}